\documentclass[runningheads]{llncs}

\usepackage{eccv}

\usepackage{eccvabbrv}

\usepackage{graphicx}
\usepackage{booktabs}
\usepackage{wrapfig}
\usepackage{capt-of}
\usepackage{multirow}
\usepackage{array}

\renewcommand{\thefootnote}{\fnsymbol{footnote}}

\usepackage[accsupp]{axessibility}  % Improves PDF readability for those with disabilities.

\usepackage[pagebackref,breaklinks,colorlinks,citecolor={green!70!black}]{hyperref}
\usepackage{orcidlink}

\begin{document}

\title{Reinforcement Learning Friendly Vision-Language Model for Minecraft} 

\titlerunning{CLIP4MC}

\author{Haobin Jiang\inst{1}$^\star$\orcidlink{0009-0009-7114-534X} \and Junpeng Yue\inst{1}$^\star$ \and Hao Luo\inst{1}\orcidlink{0000-0003-4612-5450} \and \\ Ziluo Ding\inst{2} \and Zongqing Lu\inst{1,2}$^\dag$\orcidlink{0000-0003-3967-2704}}

\renewcommand{\thefootnote}{}
\footnotetext[2]{$^\star$ Equal contribution. $^\dag$ Corresponding author.}

\authorrunning{H.~Jiang et al.}

\institute{School of Computer Science, Peking University \and
Beijing Academy of Artificial Intelligence \\
\texttt{\{haobin.jiang,zongqing.lu\}@pku.edu.cn}}

\maketitle

\begin{abstract}
  One of the essential missions in the AI research community is to build an autonomous embodied agent that can achieve high-level performance across a wide spectrum of tasks. However, acquiring or manually designing rewards for all open-ended tasks is unrealistic. In this paper, we propose a novel cross-modal contrastive learning framework architecture, CLIP4MC, aiming to learn a reinforcement learning (RL) friendly vision-language model (VLM) that serves as an intrinsic reward function for open-ended tasks. Simply utilizing the similarity between the video snippet and the language prompt is not RL-friendly since standard VLMs may only capture the similarity at a coarse level. To achieve RL-friendliness, we incorporate the task completion degree into the VLM training objective, as this information can assist agents in distinguishing the importance between different states. Moreover, we provide neat YouTube datasets based on the large-scale YouTube database provided by MineDojo. Specifically, two rounds of filtering operations guarantee that the dataset covers enough essential information and that the video-text pair is highly correlated. Empirically, we demonstrate that the proposed method achieves better performance on RL tasks compared with baselines. The code and datasets are available at \href{https://github.com/PKU-RL/CLIP4MC}{https://github.com/PKU-RL/CLIP4MC}.

  \keywords{Dataset \and Multimodal model \and Reinforcement learning}
\end{abstract}

\section{Introduction}
\label{sec:intro}

Training reinforcement learning (RL) agents to perform complex tasks in a vision-based and open-ended world can be difficult. One main challenge is that manually specifying reward functions in all open-ended tasks is unrealistic \cite{ma2023eureka,baumli2023vision}, especially when we cannot access the internal configuration of the environment. In addition, learning a reward model from human feedback is typically expensive. To this end, MineDojo \cite{fan2022minedojo} has proposed an internet-scale, multi-modal knowledge base of YouTube videos to facilitate learning in an open-ended world. With the advent of a such large-scale database, agents are able to harvest practical knowledge encoded in large amounts of media like human beings. Moreover, a vision-language model (VLM), MineCLIP \cite{fan2022minedojo}, is proposed to utilize the internet-scale domain knowledge. In more detail, the learned correlation score between the visual observation and the language prompt can be used effectively as an open-vocabulary, massively multi-task reward function for RL training. Therefore, no further task-specific reward design is needed for open-ended tasks. 

However, the autonomous embodied agent requires the vision-language model to provide a more instructive correlation score. Given the partial observations, \eg, video snippet, and the language prompt that describes the task, the agent needs to figure out two non-trivial matters to better evaluate the current state. On the one hand, \textit{whether the target entities are present within its field of vision? } MineCLIP tried to address this question. However, the alignment of texts and videos in the YouTube database is totally a catastrophe, which impedes the learning of VLM. On the other hand, \textit{what is the relationship between each video snippet and the degree of completion of the task?} Normally, in an open-ended world, \eg Minecraft, the agent explores first, then approaches and interacts with the target object. In other words, the agent requires approaching the target object before it takes trial-and-error, even for the target that needs to be kept away. Therefore, it is reasonable to make an assumption, namely, that the higher the level of completion of the task, the closer the agent is to the targets in the video snippet. We also argue it is important to incorporate the level of completion of the task into the reward function.

In this paper, we first construct neat YouTube datasets to facilitate the learning of basic game concepts, mainly the correspondence between the videos and texts. Though a large-scale database is provided by MineDojo, it contains significant noise due to its nature as an online resource. In addition, MineDojo only claims the training dataset is randomly sampled from the database, making it hard to reproduce. To overcome the catastrophic misalignment of video-text pairs in the original database, we have done four steps of dataset processing to guarantee the dataset is clean. Firstly, transcript cleaning enhances the accuracy of transcripts and ensures they are complete sentences. Secondly, keyword filtering ensures that the content of clips is relevant to the key entities in Minecraft, thereby facilitating the learning of basic game concepts. Thirdly, video partitioning and selection can handle the issue of scene transitions and thereby mitigate interference from other extraneous information. Lastly, correlation filtering can effectively address the issue of mismatch between video clips and transcripts.

We also propose an upgraded vision-language model, CLIP4MC, to provide a more RL-friendly reward function. In RL, simply utilizing the similarity between the video snippet and the language prompt is not RL-friendly since MineCLIP tends to only capture the similarity at the entity level. In other words, VLM can hardly reflect the relationship between each video snippet and the degree of completion of the task. However, this information can better help agents distinguish the importance between similar states. To achieve this, we incorporate the degree of task completion into the VLM training objective. In more detail, CLIP has exhibited remarkable segmentation capabilities without fine-tuning. After we extend pre-trained MineCLIP with modifications inspired by MaskCLIP \cite{zhou2022extract}, it can segment the specified object from the image and label the size of the target shown in the corresponding video. Intuitively, the closer the agent is to the target, the larger the target size becomes. During the learning procedure, we dynamically control the degree of contrasting positive and negative pairs of instances based on the target size in this positive video sample. Thus, CLIP4MC can render a more RL-friendly reward signal that instructs the agent to learn tasks faster. Our proposed method is trained on our YouTube dataset and evaluated on MineDojo Programmatic tasks, including harvest, hunt, and combat tasks. Empirically, our results show that CLIP4MC can provide a more friendly reward signal for the RL training procedure.

\noindent To summarize, our contributions are as follows:
\begin{itemize}
    \item \textbf{Open-sourced datasets}: We provide two high-quality datasets. The first one undergoes data cleaning (\Cref{sec:tc,sec:kf,sec:vps}) and global-level correlation filtering (\Cref{sec:ccf}). The VLM trained on this dataset matches the performance of the officially released MineCLIP, which lacks a publicly available training set.
    \item \textbf{RL-friendly dataset}: Our second dataset further incorporates local-level correlation filtering (\Cref{sec:ccf}), making it more suited for RL. The VLM trained on this dataset outperforms that on the first dataset.
    \item \textbf{RL-friendly VLM}: To better evaluate and leverage the advantages of our more RL-friendly dataset, we introduce CLIP4MC, a novel method to train a VLM that could improve downstream RL performance (\Cref{sec:algo}).
\end{itemize}

\section{Related Work}
\label{sec:related}

\noindent\textbf{Video-Text Retrieval.} Video-text retrieval plays an essential role in multi-modal research and has been widely used in many real-world web applications. Recently, the pre-trained models have dominated this line of research with noticeable results on both zero-shot and fine-tuned retrieval. Especially, BERT \cite{DevlinCLT19}, ViT \cite{dosovitskiy2021an}, and CLIP \cite{RadfordKHRGASAM21}, are used as the backbones to extract the text or video embedding. The cross-modal embeddings are then matched with specific fusion networks to find the correct video-text pair.

In more detail, CLIP4Clip \cite{LuoJZCLDL22} proposes three different similarity modules to calculate the correlation between video and text embeddings. HiT \cite{Liu0QCDW21} performs hierarchical matching at two different levels, \ie semantic level and feature level. Note that semantic level and feature level features are from the transformer network's higher and lower feature layers, respectively. Frozen \cite{BainNVZ21} proposes a dual encoder architecture that utilizes the flexibility of a transformer visual encoder to train from images or video clips with text captions. Moreover, MDMMT \cite{DzabraevKKP21} adopts several pre-training models as encoders and it shows the CLIP-based model performs the best. Therefore, our model follows this line of research by using the pre-trained model, CLIP \cite{RadfordKHRGASAM21}, to extract the feature embeddings. 

\noindent\textbf{Minecraft for AI Research.} 
As an open-ended video game with an egocentric vision, Minecraft is a noticeable and important domain in RL due to the nature of the sparse reward, large exploration space, and long-term episodes. Since the release of the Malmo simulator \cite{johnson2016malmo} and later the MineDojo simulator \cite{fan2022minedojo}, various methods have attempted to train agents to complete tasks in Minecraft \cite{tessler2017deep,shu2018hierarchical,guss2019neurips,lin2021juewu,abs-2301-04104}. Approaches such as model-based RL, hierarchical RL, goal-based RL, and reward shaping have been adopted to alleviate the sparse reward and exploration difficulty for the agent in this environment.

Recently, with the development of large language models (LLM) like GPT-4 \cite{openai2023gpt4}, a series of methods leveraging LLMs for high-level planning in Minecraft have been proposed \cite{abs-2202-10583,nottingham2023embodied,yuan2023plan4mc,wang2023voyager,zhu2023ghost}. These methods have demonstrated remarkable capabilities in guiding the agent to complete multiple complicated, long-horizon tasks, such as mining diamonds. These LLMs play a crucial role in decision-making, determining the sequence of basic skills required to accomplish specific tasks. Their effectiveness is due to their extensive knowledge about Minecraft, learned from the Internet, and their ability to reflect on real-time feedback from the game environment.

In addition to the use of LLMs, recent research attempts to incorporate Internet visual data into basic skill learning in Minecraft, beyond the traditional RL methods. MineRL \cite{GussHTWCVS19} collected 60M player demonstrations with action labels, motivating some methods \cite{ShahWWMKGWWPMGF21,abs-2202-10583} based on behavior cloning. As well-labeled data is limited in quantity, MineCLIP \cite{fan2022minedojo} instead uses over 730K narrated Minecraft videos without action labels from YouTube. It aims to learn a vision-language model providing auxiliary reward signals, utilizing the vast and diverse data available on the Internet. Different from MineCLIP, VPT \cite{abs-2206-11795} uses action-labeled data to train an inverse dynamic model to label 70K hours of Internet videos and then conduct behavior cloning.

Unlike the existing approaches, which incorporate the human experience and require a large number of demonstrations with action labels to train the agent, our work follows the line of MineCLIP \cite{fan2022minedojo} and focuses on only using the data without action labels to assist agent learning in Minecraft, which is more friendly with data collection and has the potential to scale in the future.

\section{Background}
\label{sec:background}

\noindent\textbf{MineDojo tasks.}
MineDojo \cite{fan2022minedojo} provides thousands of benchmark tasks, which can be used to develop generally capable agents in Minecraft. These tasks can be divided into two categories, Programmatic and Creative tasks. The former has ground-truth simulator states to assess whether the task has been completed. The latter, however, do not have well-defined success criteria and tend to be more open-ended, but have to be evaluated by humans. 

We mainly focus on Programmatic tasks since they can be automatically assessed. Specifically, MineDojo provides 4 categories of programmatic tasks, including Harvest, Combat, Survival, and Tech Tree, with 1581 template-generated natural language goals to evaluate the agent’s different capabilities. Among these tasks, Survival and Tech Tree tasks are harder than Harvest and Combat tasks. Currently, MineCLIP \cite{fan2022minedojo} only expresses promising potential in some Harvest and Combat tasks. Harvest means finding, obtaining, cultivating, or manufacturing hundreds of materials and objects. Combat means fighting various monsters and creatures that require fast reflexes and martial skills. 

\noindent\textbf{POMDP.}
We model the programmatic task as a partially observable Markov decision process (POMDP) \cite{kaelbling1998planning}. At each timestep $t$, the agent obtains the partial observation $o_t$ from the global state \(s_t\) and a language prompt \(G\), takes action \(a_{t}\) following its policy \(\pi(a_{t}|o_{t})\), and receives a reward \(r_{t} = \Phi(V_{t},G)\), where $V_t$ is the fixed-length sequence of observations till $t$ (thus a video snippet) and \(\Phi\) maps \(V_t\) and \(G\) to a scalar value. Then the environment transitions to the next state \(s_{t+1}\) given the current state and action according to transition probability function \(\mathcal{T}(s_{t+1}|s_t,a_t)\). The agent aims to maximize the expected return \(R=\mathbb{E}_\pi\sum_{t=1}^{T}\gamma^{t-1}r_{t}\), where \(\gamma\) is the discount factor and \(T\) is the episode time horizon. 

\section{YouTube Dataset}
\label{sec:dataset}

\begin{figure}[t]
    \centering
    \includegraphics[width=.95\linewidth]{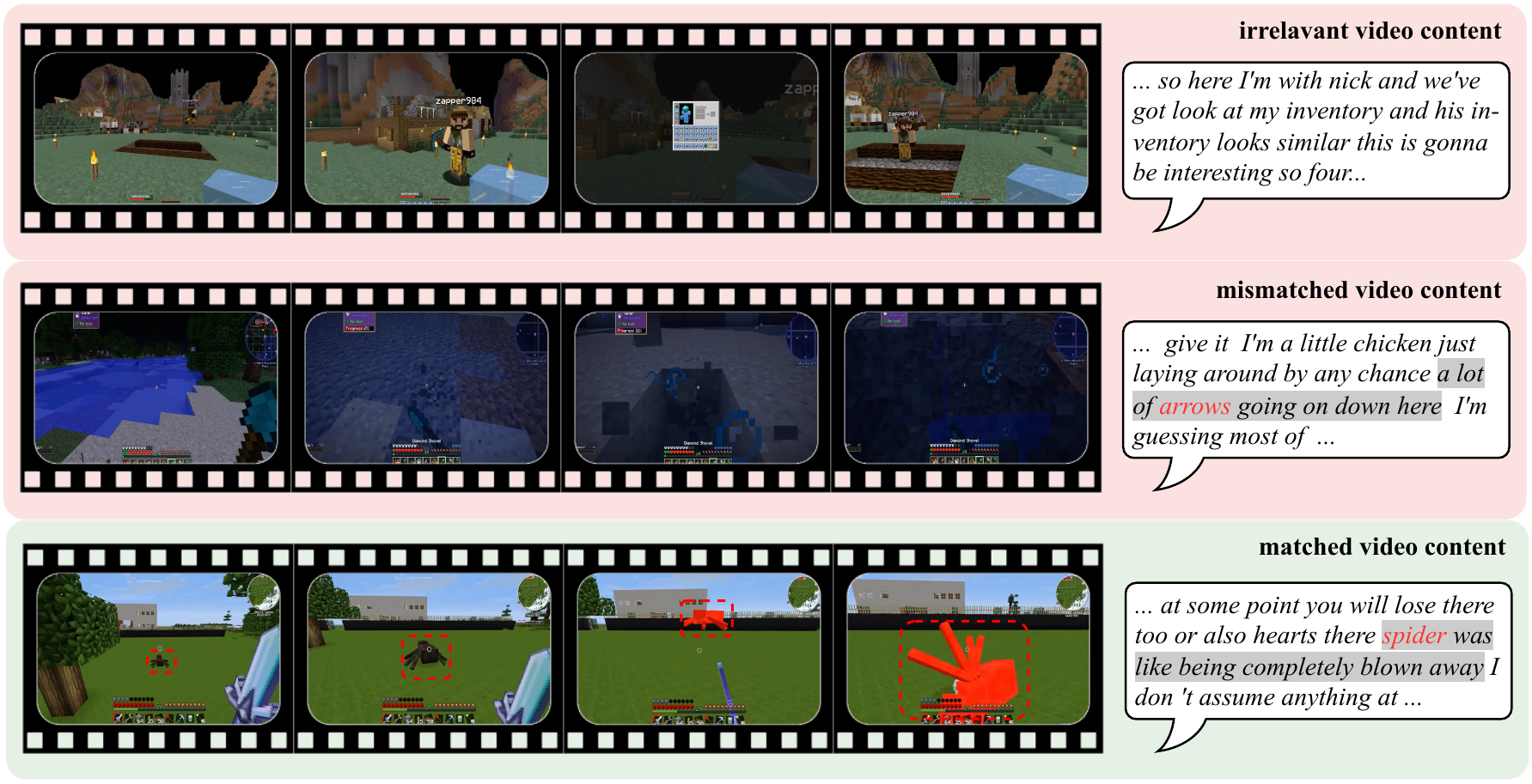}
    \caption{Illustration of the YouTube video database. The screenshots of video clips are on the left and key entities are circled in red. The corresponding transcript clips are on the right and key entities are marked in red. We give examples of irrelevant, mismatched, and matched video content in the YouTube video database.}
    \label{fig:dataset}
    \vspace{-2mm}
\end{figure}

The Internet is rich in Minecraft-related data, containing a wealth of weakly labeled or even unlabeled Minecraft knowledge, including crucial entities, plausible actions, and common-sense event processes. With these multi-modal data, it is possible to create dense language-conditioned rewards, making open-ended long-term task learning feasible. MineDojo \cite{fan2022minedojo} collected over 730K YouTube videos and their corresponding transcripts, totaling 33 years of video content and 2.2B words in transcripts. Around 640K video clips are selected using keyword-based filtering from these videos, and these clips are used to train MineCLIP. However, MineDojo open-sourced a 13.8M dataset, and the 640K video clips were randomly sampled in their paper. However, as illustrated in \Cref{fig:dataset}, most videos feature irrelevant game content that is not conducive to learning basic game concepts. Meanwhile, the alignment between the transcripts and videos may not always be precise, leading to temporal or content discrepancies that could hinder the learning of retrieval and RL tasks. Given the low quality of the data, we found it necessary to provide a neat 640k dataset to the community for the training usage. To address these issues, we adopt the following few steps to acquire a clean dataset with high quality.

\subsection{Transcript Cleaning}
\label{sec:tc}

Downloading YouTube's automatic transcripts directly can lead to several issues. Firstly, transcript blocks may overlap, resulting in overlapping timestamps in the transcript. Additionally, the caption quality is typically mediocre, with a higher occurrence of misidentifications, especially in non-English videos. Moreover, the transcript lacks punctuation, making it less friendly for understanding semantics. Based on the aforementioned issues, we implement a pipeline to construct high-quality transcripts as follows: (1) Extract audio from the videos and use Whisper \cite{radford2022whisper} to obtain high-quality, temporally non-overlapping transcripts. (2) Employ FullStop \cite{guhr-EtAl:2021:fullstop} to generate punctuation, resulting in complete sentences of 10-35 words in length.

\subsection{Keyword Filtering}
\label{sec:kf}

Following MineDojo \cite{fan2022minedojo}, we also implement keyword-based filtering to ensure that the content in our dataset is pertinent to the key entities in Minecraft, thereby facilitating the learning of basic game concepts. As essential components of Minecraft, the key entities, such as stones, trees, and sheep, are common across multiple tasks in MineDojo and videos from YouTube. Specifically, we identify entity keywords in the transcripts using a keyword list from MineDojo and extract transcript clips formed into sentences to encompass as many keywords as possible. These extracted transcript clips then serve as the textual component of our dataset, determining the location of corresponding video clips.

\subsection{Video Partition and Selection}
\label{sec:vps}

After completing the previous two steps, we obtain transcript clips relevant to the keywords. For each transcript clip, we calculate the central timestamp that corresponds to the clip based on the transcript timestamps and then use this central timestamp to extract a video clip with a duration of $D$ seconds from the video. This process allows us to obtain temporally-aligned video clips. However, the video clips obtained in this manner still exhibit some issues, including scene transitions and discrepancies between video content and transcripts. We handle the former problem through video partition and filtering, while the latter problem is addressed in \Cref{sec:ccf}.

Owing to the informal nature of YouTube content, there is often a lack of semantic congruence between the video clips and their corresponding transcriptions, as noted in VideoCLIP \cite{xu2021videoclip}. Moreover, video clips frequently contain a few different behaviors which cause scene transitions, \eg Chopping down the tree first, then suddenly switching the inventory bar. Since some scene transitions lead to irrelevant information, we partition the video content into several semantically coherent segments based on the semantic structure of the video. Then we select the segment that aligns best with the transcript.

To achieve semantic partition of video content, we employ the Bellman $k$-segmentation algorithm \cite{haiminen2008algorithms}. This algorithm divides a sequence of data points into $k$ distinct and constant-line segments, providing a piece-wise constant approximation of the data. To process the video, we first use the officially released MineCLIP \cite{fan2022minedojo} video encoder to obtain the embedding of each frame, since MineCLIP can capture video semantics to some extent. Subsequently, we partition these embeddings into $k$ segments and select the segment with the highest similarity score, as calculated in MineCLIP.

\subsection{Correlation Filtering}
\label{sec:ccf}

\begin{figure}[t]
    \centering
    \includegraphics[width=0.7\columnwidth]{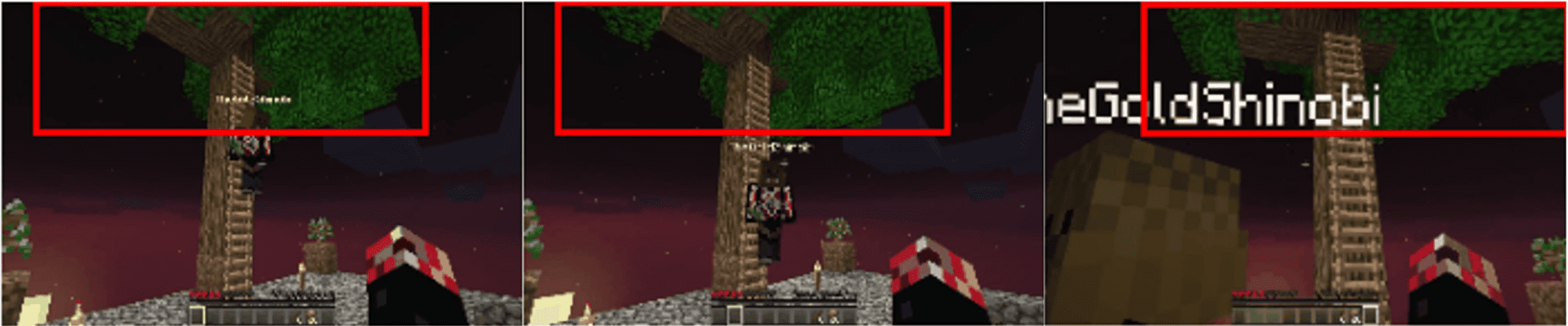}
    \includegraphics[width=0.7\columnwidth]{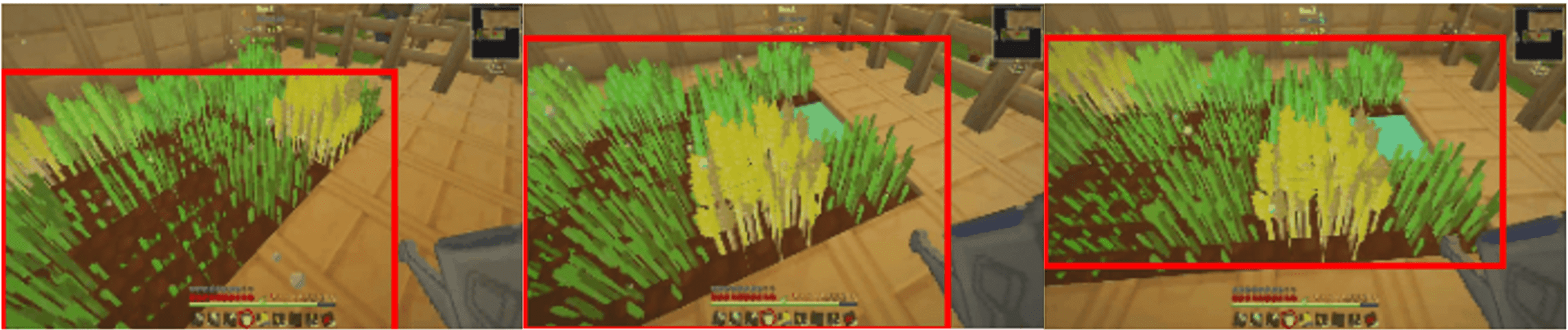}
    \includegraphics[width=0.7\columnwidth]{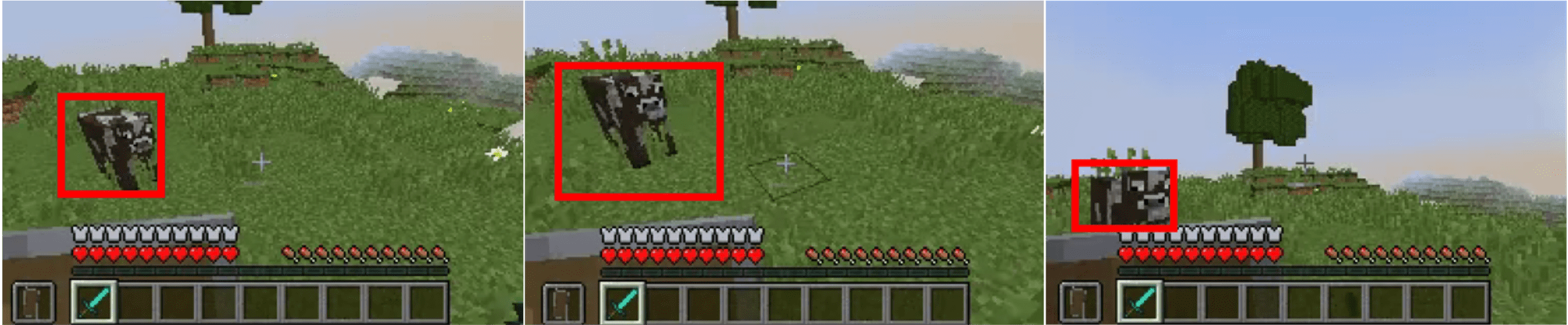}
    \caption{Examples of how we estimate the size of key entities in video frames. Red bounding boxes are generated by our modified MineCLIP visual encoder, following the approach proposed in MaskCLIP \cite{zhou2022extract}. These boxes are then used to calculate the size.}
    \label{fig:bbox}
    \vspace{-2mm}
\end{figure}

We employ correlation filtering techniques to address disparities that persist between video content and transcripts. The correlation filtering is done at two levels, the global and the local levels. From the global level, we calculate the cosine similarities between video embeddings and text embeddings via the original MineCLIP and then select clips based on these similarities.

Recent research \cite{zhou2022extract,li2023clip} has demonstrated that CLIP \cite{RadfordKHRGASAM21}, though trained on whole images, can generate meaningful local features. Inspired by this and following MaskCLIP \cite{zhou2022extract}, we make modifications to the original MineCLIP visual encoder, empowering it with the ability to estimate and label the size of the key entity, which is mentioned in the corresponding transcript, in each frame of a video clip without fine-tuning. Then the correlations between video clips and transcripts are calculated as the summation of the sizes across all frames in each video clip. We consider this correlation at the local level.

Based on these two criteria, we select the top $k\%$ of clips, resulting in the final training set. We provide an elaboration on our implementation in \Cref{app:dataset} and illustrate some examples in \Cref{fig:bbox}.

\vspace{1mm}
\noindent The aforementioned four-step approach creates a dataset consisting of 640K video-text clip pairs, with an additional 4K pairs extracted for validation of video-text retrieval. Regarding the constants of the approach, $D$ and $k$ are set to 16 and 50, respectively. Therefore, our dataset comprises videos with a total duration of one week and approximately 0.16B words, significantly smaller in scale compared to the original low-quality 13.8M dataset. Importantly, we will open-source our entire YouTube dataset, serving as an upgraded version of the unreleased MineCLIP training data.

\section{CLIP4MC}
\label{sec:clip4mc}

Given a video snippet \(V\) and a language prompt \(G\), the vision-language model outputs a correlation score, \(C\), that measures the similarity between the video snippet and the language prompt. Ideally, if the agent performs behaviors following the description of the language prompt, the vision-language model will generate a higher correlation score, leading to a higher reward. Otherwise, the agent will be given a lower reward.

\subsection{Architecture}

We follow the same architecture design as MineCLIP \cite{fan2022minedojo}, including a video encoder, a text encoder, and a similarity calculator. All the video frames first go through the spatial transformer to obtain a sequence of frame features. The temporal transformer is then utilized to summarize the sequence of frame features into a single video embedding. An adapter further processes the video embedding for better features. Refer to \Cref{app:arch} for more details about the architecture. 

Empirically, we found that the video encoder (essentially MineCLIP) can provide a bond between the entities and the language prompts and give a similar high reward as long as the target entities are present in the video frames, similar to observations in \cite{cai2023open}. However, such a reward is not instructive enough for RL tasks since it does not reflect the behavioral trends of agents.

\subsection{Contrastive Training}
\label{sec:algo}

We aim to minimize the sum of the multi-modal contrastive losses \cite{oord2018representation}, including video-to-text, and text-to-video:
\begin{equation}
\begin{aligned}
\mathcal{L}  = & -\sum_{(z_v,z_m,z_t)\in \mathcal{B}}\Big(\log {\rm NCE}(z_v,z_t)+\log {\rm NCE}(z_t,z_v)  \Big),
\end{aligned}
\end{equation}
where \(\mathcal{B}\) is the batch containing sampled video-text pairs, and \({\rm NCE}(\cdot,\cdot)\) is the contrastive loss that calculates the similarity of two inputs. To illustrate, the video-to-text contrastive loss is given by 
\begin{equation}
    {\rm NCE}(z_v,z_t) = \frac{\exp(z_v \cdot z_t^{+}/\lambda)}{\sum_{z \in \{z_t^{+},z_t^{-}\}} \exp(z_v \cdot z/\lambda)},
\end{equation}
where \(\lambda\) is a temperature hyperparameter, \(z_t^{+} \)
is the positive text embedding matching with the video embedding \(z_v\), and \(\{z_t^{-}\}\) are negative text embeddings that are implicitly formed by other text clips in the training batch. Contrastive loss of text-to-video is defined in the same way.

In addition, we also need to incorporate the degree of task completion into the training objective. Specifically, we hope the similarity score between video clip and text embeddings could reflect the task completion degree, \ie, higher completion leads to higher similarity. The sizes of key entities provided by the local correlation filtering serve as a surrogate for the task completion degree. Initially, we try to dynamically adjust the weight of the positive pairs based on the size of the target object. However, it did not work. Essentially, positive and negative pairs will still be separated, even with smaller sample weights. 

\begin{figure}[t]
    \centering
    \includegraphics[width=0.69\columnwidth]{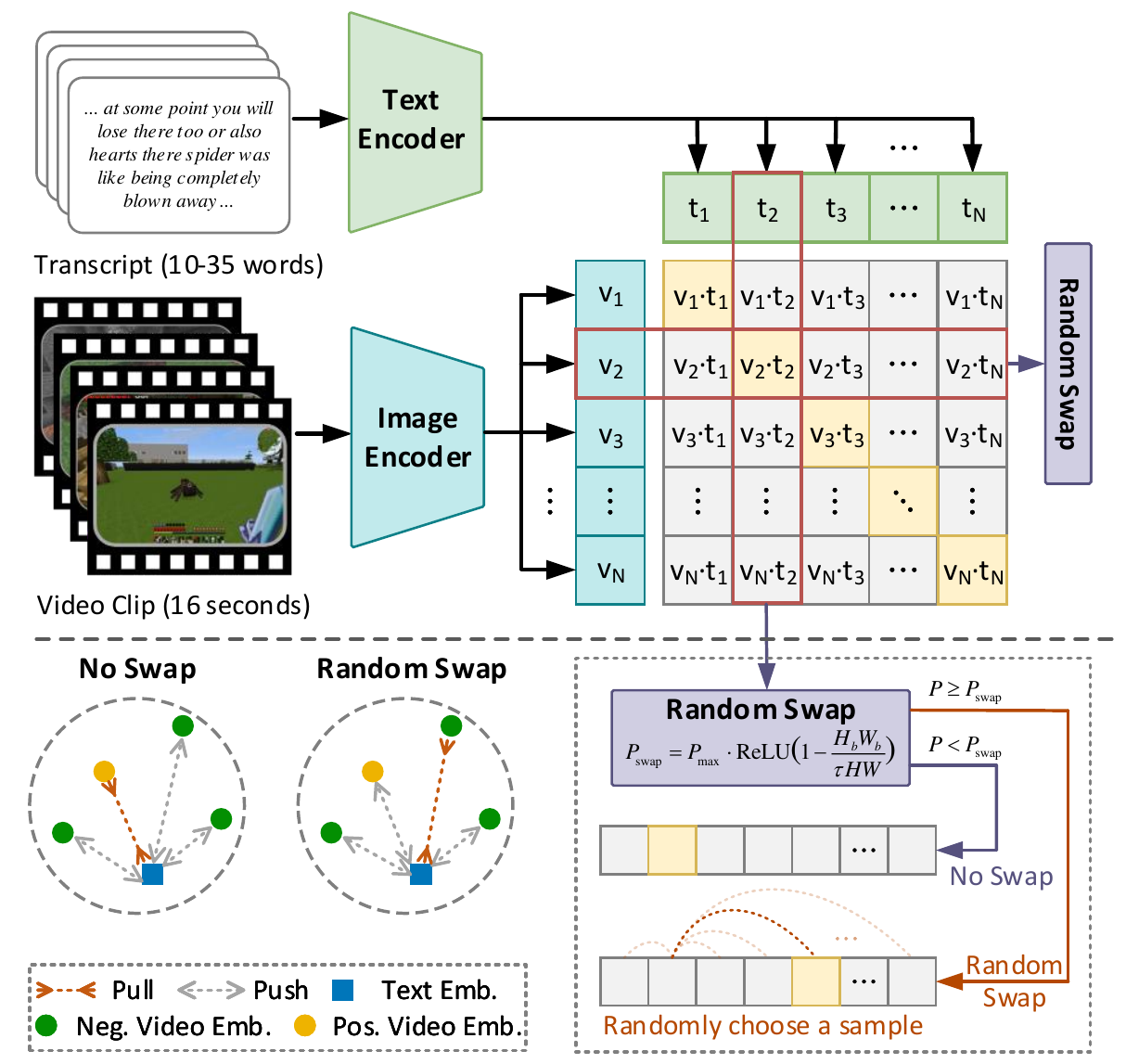}
    \caption{Illustration of CLIP4MC training. The upper part shows the concept of contrastive learning, while the lower part explains the swapping operation.}
    \label{fig:swap}
    \vspace{-2mm}
\end{figure}

Therefore, we instead forcibly change the labels of the positive and negative video samples, as illustrated in \Cref{fig:swap}, thereby reducing the model's confidence in certain positive pairs. Specifically, during training, a positive video sample is swapped with a random negative video sample from the \(\mathcal{B}\) based on probability $P_{\rm swap}$. This swap only occurs when the labeled target size $H_b W_b$ 
in the positive video sample is below a certain threshold $\tau$, and the smaller the size, the greater the probability,
\begin{equation}
P_{\rm swap} = P_{\rm max} *{\rm ReLU} \left( 1- \frac{H_b W_b}{\tau HW}\right),
\label{equ:clip4mc}
\end{equation}
where $HW$ represents the size of the image. 

Since contrastive learning brings positive samples closer and pushes negative samples away, converting a positive pair to a negative one with a certain probability, \ie, the aforementioned random swap, during training will lower the similarity score. As the decrease in similarity is directly proportional to the swapping probability, we set $P_{\rm swap}$ to be inversely related to task completion, ensuing similarity score increases as completion increases.

In addition, the upper limit $P_{\rm max}$ in \Cref{equ:clip4mc} is set to 0.5, as the swap should not disrupt severely the normal distinction between positive and negative pairs in most cases. This constraint enables CLIP4MC to preserve the understanding capability of MineCLIP for general behaviors that may lack explicit target entities in Minecraft, as discussed in \Cref{app:creative}.

\subsection{RL Training} 
\label{sec:reward}

For RL training, the first step is reward generation. At timestep \(t\), we concatenate the agent’s latest 16 egocentric RGB frames in a temporal window to form a video snippet, \(V_t\). CLIP4MC outputs the probability \(P_{G,t} \) that calculates the similarity of \(V_t\) to the task prompt, \(G\), against all other negative prompts. To compute the reward, we further process the raw probability as previous work \cite{fan2022minedojo} \(r_t = \max ( P_{G,t} - 1 / N_t,0) \), where \(N_t\) is the number of prompts passed to CLIP4MC. Note that CLIP4MC can handle unseen language prompts without any further fine-tuning due to the open-vocabulary ability of CLIP \cite{RadfordKHRGASAM21}. 

The ultimate goal is to train a policy network that takes as input raw pixels and other structural data and outputs discrete actions to accomplish the task that is described by the language prompt, \(G\). We use PPO \cite{schulman2017proximal} as our RL training backbone and the policy is trained on the CLIP4MC reward together with the sparse task reward if any. The policy input contains several modality-specific components and more details can be found in \Cref{app:env}.

\section{Experiments}
\label{sec:exp}

In this section, we comprehensively evaluate and analyze our proposed model CLIP4MC, utilizing the open-ended platform MineDojo \cite{fan2022minedojo}, which comprises thousands of diverse, open-ended Minecraft tasks designed for embodied agents. 

We compare CLIP4MC against two baselines: (1) \textbf{MineCLIP [official]}, the officially released MineCLIP model \cite{fan2022minedojo}. (2) \textbf{MineCLIP [ours]}, using the same architecture as MineCLIP [official] but trained on our cleaned YouTube dataset. It also serves as the ablation of CLIP4MC without the swap operation. We train CLIP4MC and MineCLIP [ours] for 20 epochs and select the models with the highest performance on RL tasks. Please refer to \Cref{app:train} for more training details. All results are presented in terms of the mean and standard error of four runs with different random seeds.

\subsection{Environment Settings}

We conduct experiments on eight Programmatic tasks, comprising two harvest tasks: \textit{milk a cow} and \textit{shear wool}, two combat tasks: \textit{combat a spider} and \textit{combat a zombie}, and four hunt tasks: \textit{hunt a cow}, \textit{hunt a sheep}, \textit{hunt a pig}, and \textit{hunt a chicken}. These tasks are all built-in tasks in the MineDojo benchmark.

\vspace{1mm}
\noindent \textbf{Harvest.} \textit{Milk a cow} requires the agent to obtain milk from a cow with an empty bucket. Similarly, \textit{shear wool} requires the agent to obtain wool from a sheep with shears. A harvest task is terminated and considered completed when the target item is obtained by the agent with a specified quantity. The prompts used to calculate the reward is ``obtain milk from a cow in plains with an empty bucket''; for \textit{shear wool}, it is ``shear a sheep in plains with shears''.

\vspace{1mm}
\noindent \textbf{Combat.} In these tasks, target animals, spiders, and zombies, are hostile and will actively approach and attack the agent. The agent's goal is to fight and kill the target animal. The prompt for each combat task is ``combat a spider/zombie in plains with a diamond sword''.

\vspace{1mm}
\noindent \textbf{Hunt.} Hunt tasks consist of \textit{hunt a cow}, \textit{hunt a sheep}, \textit{hunt a pig}, and \textit{hunt a sheep}. For each task, the agent's goal is to kill the target animal as indicated in the task name. Different from combat tasks, the target animals will flee from the agent after being attacked. Therefore, these tasks require the agent to keep chasing and attacking the target, making them challenging. As noted in \cite{cai2023open}, the original MineCLIP reward fails in these tasks since it cannot consistently increase when the agent approaches the agent. This observation aligns with our assertion that the original MineCLIP model can hardly capture the degree of task completion. The prompt for each task is ``hunt a \{target\} in plains with a diamond sword'' where \{target\} is replaced with the corresponding animal name. 

\vspace{1mm}
More elaborated introduction of the Minecraft environment and settings of these tasks are available in \Cref{app:env}. To guarantee a fair comparison, we adopt the same RL hyperparameters for all models and tasks. These hyperparameters are listed in \Cref{app:rl}.

\subsection{RL Results}

\begin{table}[t]
    \centering
    \renewcommand{\arraystretch}{1.2}
    \caption{Success rates (\%) of RL trained with rewards provided by different models on eight Minecraft tasks. Each mean and standard error of success rates are calculated on four models after training 1e6 environment steps with different random seeds.}
    \begin{footnotesize}
    \begin{tabular}{p{0pt} c p{0pt} >{\centering\arraybackslash}p{50pt} p{0pt} >{\centering\arraybackslash}p{52pt} p{0pt} >{\centering\arraybackslash}p{64pt} p{0pt} >{\centering\arraybackslash}p{67pt} p{0pt}} \toprule
     & \multirow{2}{*}{\textbf{Models}} & & \multicolumn{3}{c}{\textbf{Harvest}} & & \multicolumn{3}{c}{\textbf{Combat}} & \\
     \cline{4-6} \cline{8-10}
     \rule{0pt}{2.8ex} & & & \textit{milk a cow} & & \textit{shear wool} & & \textit{combat a spider} & & \textit{combat a zombie} & \\ \midrule
     & CLIP4MC    & & \textbf{84.5±2.0}  & & \textbf{74.6±2.1} & & \textbf{85.8±0.9} & & \textbf{70.4±8.3} & \\
     & MineCLIP[ours]   & & 84.4±1.1 & & 71.6±3.5   & & 75.4±10.1  & & 63.6±8.7 & \\
     & MineCLIP[official]    & & 84.1±0.5  & & 73.2±1.8 & & 82.7±2.5 & & 57.4±3.7 & \\
    \bottomrule
    \end{tabular}
    \par\vspace{3mm}
    \begin{tabular}{p{0pt} c p{0pt} >{\centering\arraybackslash}p{50pt} p{0pt} >{\centering\arraybackslash}p{52pt} p{0pt} >{\centering\arraybackslash}p{64pt} p{0pt} >{\centering\arraybackslash}p{67pt} p{0pt}} \toprule
     & \multirow{2}{*}{\textbf{Models}} & & \multicolumn{7}{c}{\textbf{Hunt}} & \\
     \cline{4-10}
     \rule{0pt}{2.8ex} & & & \textit{hunt a cow} & & \textit{hunt a sheep} & & \textit{hunt a pig} & & \textit{hunt a chicken} & \\ \midrule
     & CLIP4MC    & & \textbf{39.8±2.5}  & & \textbf{45.9±7.2} & & \textbf{30.6±8.4} & & \textbf{26.1±3.5} & \\
     & MineCLIP[ours]   & & 17.3±10.6 & & 33.0±18.1 & & 14.1±10.6 & & 15.3±10.6 & \\
     & MineCLIP[official]    & & 11.6±11.1  & & 28.5±16.7 & & 1.5±0.6 & & 0.0±0.0 & \\
    \bottomrule
    \end{tabular}
    \end{footnotesize}
\label{tab:mt_combat}
\vspace{-2mm}
\end{table}

Through our evaluation of CLIP4MC, MineCLIP [ours], and MineCLIP [official] across eight Minecraft tasks, we want to answer two key questions:

\vspace{1mm}
\noindent (1) Whether the YouTube dataset we constructed enables MineCLIP, when trained on it, to provide a more effective reward signal for task learning?

\vspace{1mm}
\noindent (2) Whether our upgraded model, CLIP4MC, based on our YouTube dataset, offers a reward signal that is further friendly for the RL training procedure?

\vspace{1mm}
\noindent These questions are central to verifying the effectiveness of our dataset and CLIP4MC model in Minecraft. Table \ref{tab:mt_combat} shows the success rates of all methods on the eight Minecraft tasks.

It is noticeable that, in four hunt tasks, three models demonstrate varying performance on RL. Firstly, MineCLIP [ours] consistently achieves better results compared to MineCLIP [official] across all hunt tasks. The superior performance of MineCLIP [ours] provides a positive answer to our first question, suggesting that the YouTube dataset we construct indeed enhances the effectiveness of the reward signal in MineCLIP for task learning. \textit{Note that this is the dataset we plan to release for better training the VLM model for RL tasks on Minecraft.} Secondly, CLIP4MC shows significantly higher success rates on hunt tasks compared to both MineCLIP [ours] and MineCLIP [official], meaning that the answer to the second question is also positive. As CLIP4MC provides a reward signal taking into account a surrogate for the degree of task completion, \ie, the size of the target object in our implementation, it becomes more RL-friendly in these challenging tasks.

In addition, we notice a practical example of misalignment in the officially released MineCLIP model. Specifically, we observe that in \textit{hunt a chicken}, the agent trained with MineCLIP [official] tends to keep looking at the sky, indicating such behavior can provide a high intrinsic reward. This phenomenon suggests that the officially released MineCLIP indeed suffers from the misalignment problem in the YouTube dataset. In contrast, our MineCLIP [official] shows promising behaviors on this task, demonstrating that our dataset processing improves the alignment between the transcripts and videos.

In contrast to hunt tasks, these methods perform comparably on harvest tasks and combat tasks, while CLIP4MC still shows marginal advantage over the other two methods. This finding aligns with the results reported in \cite{fan2022minedojo}, where the MineCLIP reward already achieves saturated performance on these \textit{easy} tasks. Therefore, further enhancements in the reward signal, like MineCLIP [ours] and CLIP4MC, do not significantly improve performance. Unlike hunt tasks, neither harvest nor combat tasks require the agent to take multiple rounds of chasing. This result does not detract from the superiority of CLIP4MC, evidenced by its performance on hunt tasks, which are considered more difficult \cite{abs-2206-11795}.

\subsection{Reward Analysis}

\begin{figure}[t]
   \centering
   \begin{subfigure}{0.23\textwidth}
		\centering
		\includegraphics[width=\textwidth]{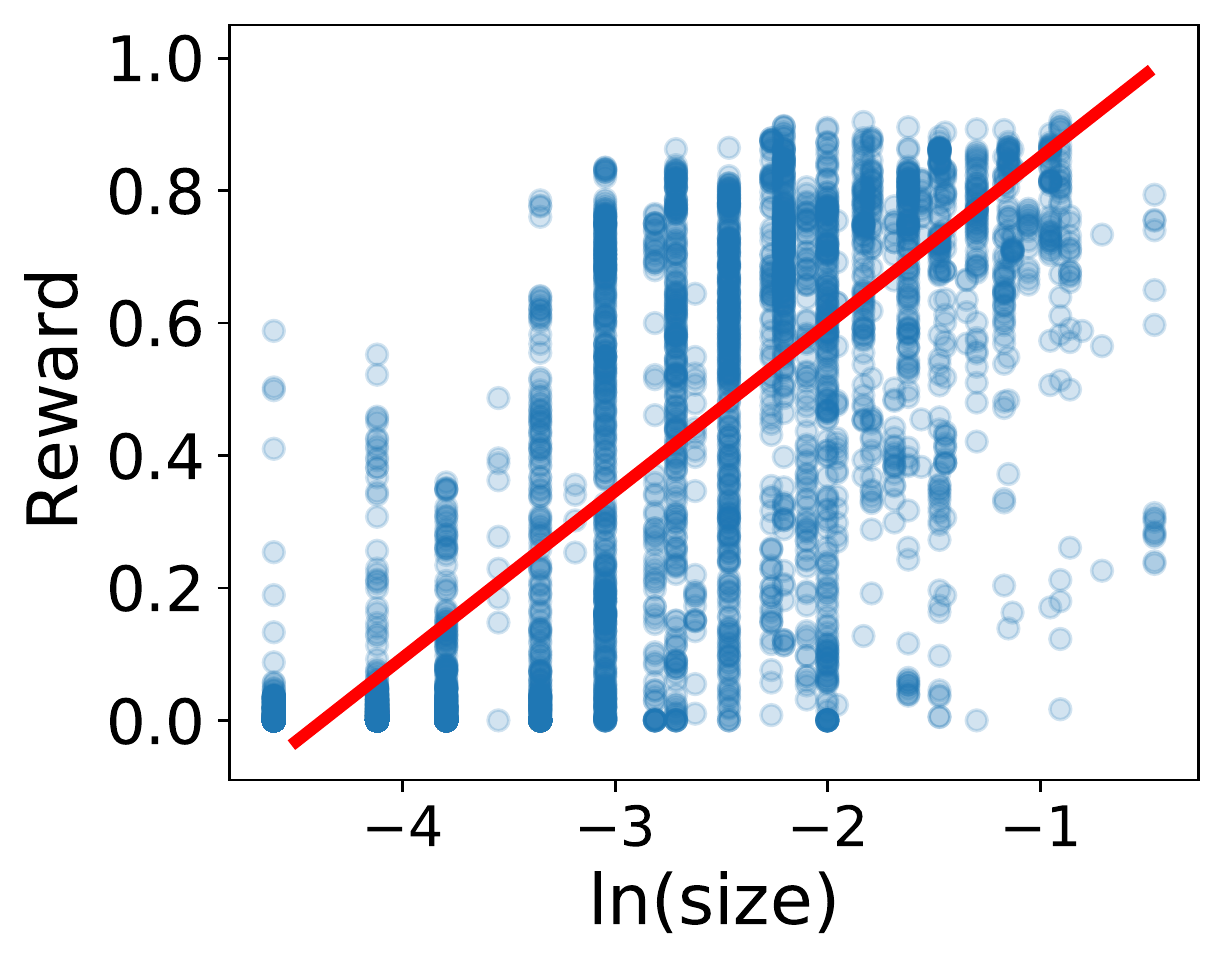}
		\caption{CLIP4MC}
    \end{subfigure}
    \quad
    \begin{subfigure}{0.23\textwidth}
		\centering
		\includegraphics[width=\textwidth]{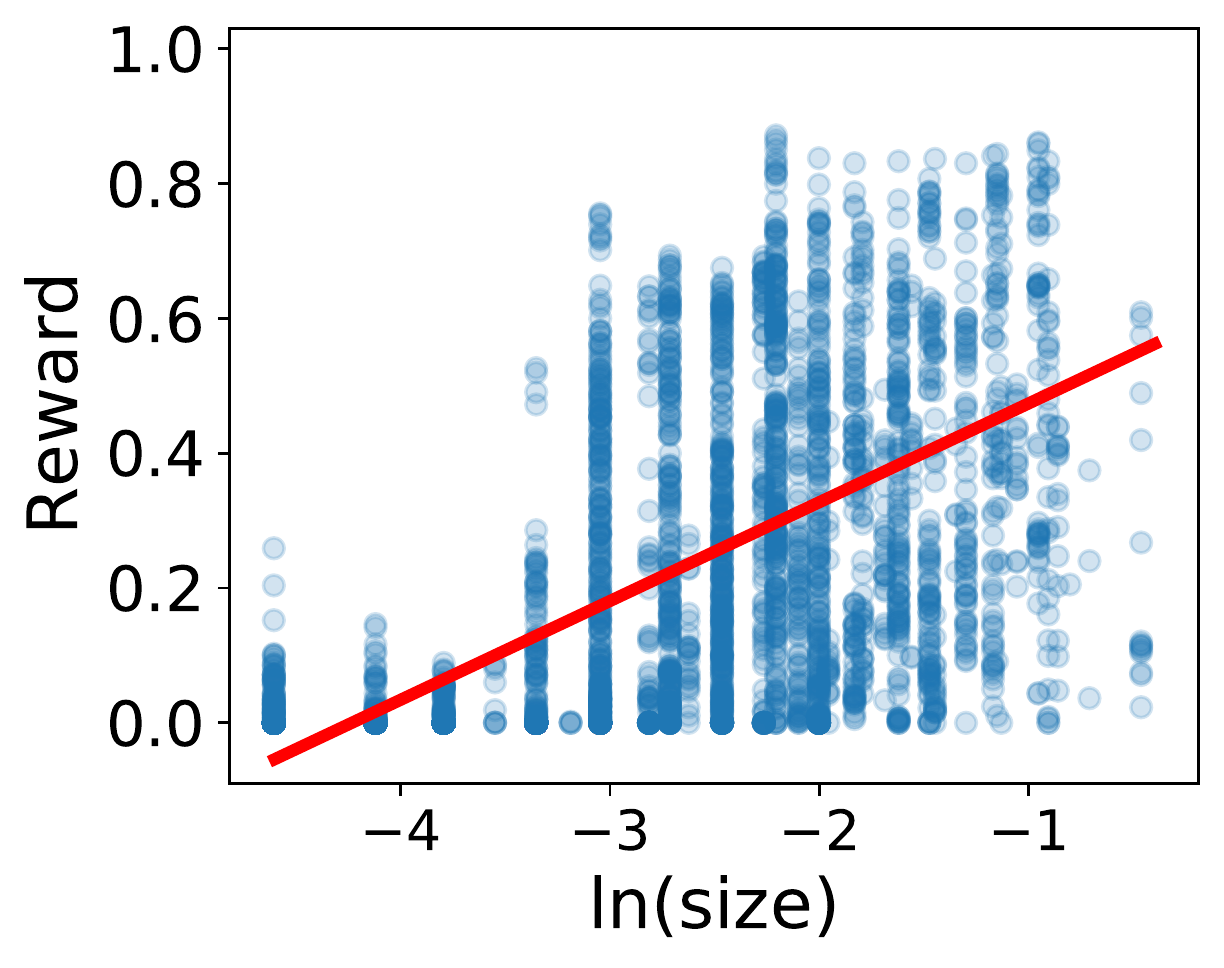}
		\caption{MineCLIP [ours]}
    \end{subfigure}
    \quad
    \begin{subfigure}{0.23\textwidth}
		\centering
		\includegraphics[width=\textwidth]{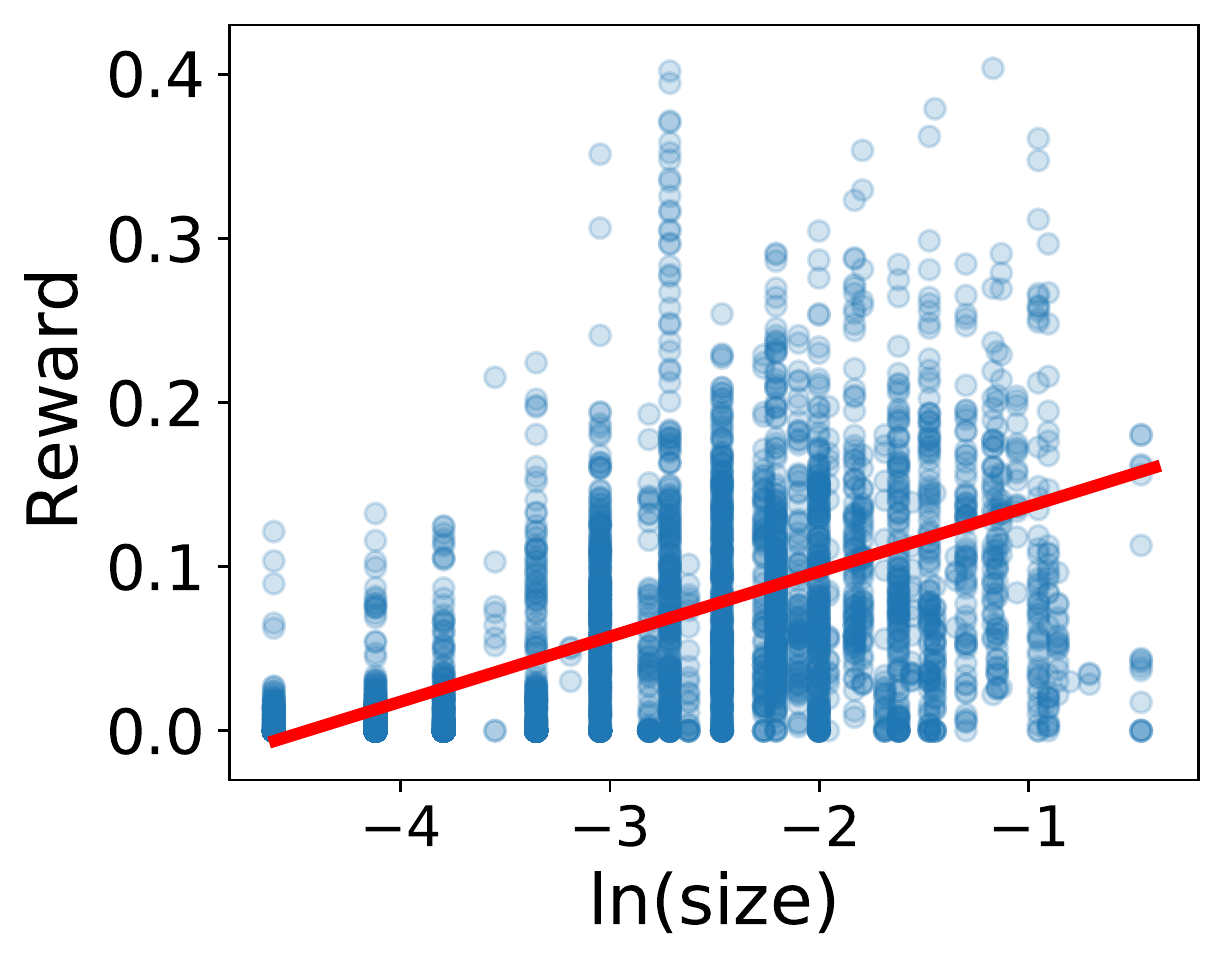}
		\caption{MineCLIP [official]}
    \end{subfigure}
    \caption{Scatter plots illustrating the relationship between the entity size and the intrinsic reward. The red line indicates a linear fit to the data.}
    \label{fig:rew}
    \vspace{-2mm}
\end{figure}

To quantitatively verify that CLIP4MC captures the size of the target entity specified in the language prompt, we collect 5000 steps in task \textit{hunt a cow} and apply the method described in \Cref{app:dataset} to estimate the maximal size of the cow in consecutive 16 frames. Then we use CLIP4MC, MineCLIP [ours], and MineCLIP [official] to calculate intrinsic rewards respectively. Before calculating the correlation between the size and intrinsic rewards, we transform the size value using $f(x) =\ln{(x + e^{-2})}$, focusing on smaller values. The relationship between the transformed size and intrinsic rewards is visualized in \Cref{fig:rew}. The corresponding Pearson correlation coefficients from left to right are 0.81, 0.66, and 0.62, indicating that CLIP4MC reward has a higher correlation with the size, especially when it is relatively small. This is crucial in RL, as the agent needs dense and distinguishing reward signals to guide the learning process, particularly when the target is distant. Such characteristic is the essential benefit of CLIP4MC in RL.

\subsection{Ablation on Dataset Filtering}

\begin{table}[t]
    \centering
    \caption{Success rates (\%) of RL trained with rewards provided by different models on three Minecraft tasks. Each mean and standard error of success rates are calculated on four models after training 5e5 and 1e6 environment steps with different random seeds.}
    \renewcommand{\arraystretch}{1.2}
    \begin{footnotesize}
    \begin{tabular}{p{2pt} c p{2pt} c p{2pt} c p{2pt} c p{2pt} c p{2pt} c p{2pt} c p{2pt} c p{2pt}} \toprule
     & \multirow{2}{*}{\textbf{Models}} & & \multicolumn{3}{c}{\textit{milk a cow}} & & \multicolumn{3}{c}{\textit{shear wool}} & & \multicolumn{3}{c}{\textit{hunt a cow}} \\
     \cline{4-6} \cline{8-10} \cline{12-15}
     \rule{0pt}{2.8ex} & & & 5e5 & & 1e6 & & 5e5 & & 1e6 & & 5e5 & & 1e6 & \\ \midrule
     & ours  & & \textbf{73.1±0.9}  & & \textbf{84.4±1.1} & & 47.5±6.3 & & 71.6±3.5 & & \textbf{15.6±8.7} & & \textbf{17.3±10.6} & \\
     & w/o LCF & & 71.3±1.8 & & 80.7±1.1  & & \textbf{53.8±3.3}  & & \textbf{73.6±1.5} & & 0.7±0.7 & & 12.1±12.0 & \\
     & RS & & 65.1±3.9 & & 81.7±1.0 & & 21.7±6.5 & & 68.5±3.6 & & 1.0±0.5 & & 4.0±2.6 & \\
     & official & & 69.8±1.3 & & 84.1±0.5 & & 47.0±13.5 & & 73.2±1.8 & & 2.7±1.1 & & 11.6±11.1 \\
    \bottomrule
    \end{tabular}
    \end{footnotesize}
    \label{tab:ablation}
\end{table}

To evaluate the influence of the dataset quality on RL training and verify the effectiveness of our proposed correlation filter at the local level, we compare MineCLIP [ours] with two ablations: (1) \textbf{MineCLIP [w/o LCF]} is trained on the dataset processed with only global correlation filtering, omitting the local correlation filtering. (2) \textbf{MineCLIP [RS]} is trained on 640K video clips randomly sampled from the MineDojo released database, consistent with their stated dataset construction method \cite{fan2022minedojo}. Note that methods evaluated here \textit{do not} apply random swap introduced in \Cref{sec:algo} since these datasets cannot provide entity size. The results are presented in \Cref{tab:ablation}. Given the \emph{overall} performance across all three presented tasks, firstly, [RS] does not perfectly reproduce the performance of [official], suggesting that misalignment in the original database hinders the reproduction of [official]. Secondly, part of our proposed data filtering makes the model, [w/o LCF], comparable to [official]. In addition, with all of our proposed data filtering techniques, [ours] outperforms [official].

\subsection{Video-Text Retrieval Results}

\begin{wrapfigure}{r}{0.6\textwidth}
    \centering
    \vspace{-8mm}
    \renewcommand{\arraystretch}{1.2}
    \captionof{table}{Results of video-to-text / text-to-video retrieval on the test set. The best results are highlighted in \textbf{bold}.}
    \vspace{4mm}
    \begin{footnotesize}
    \begin{tabular}{lccc}
        \toprule
        \multicolumn{1}{c}{\textbf{Models}} & R@1 & R@5 & R@10 \\
        \midrule
         ours & 12.4/13.1 & 27.5/27.8 & 35.3/35.9 \\
        w/o LCF & \textbf{12.7/14.1} & \textbf{29.4/29.7} & \textbf{37.9/37.5} \\
        RS & 11.1/11.6 & 24.8/25.0 & 32.4/32.2 \\
        \bottomrule
    \end{tabular}
    \end{footnotesize}
    \label{tab:v2t}
\end{wrapfigure}
\Cref{tab:v2t} shows the results of video-to-text retrieval and text-to-video on the test set with the same MineCLIP model for a fair comparison. We train these models for 20 epochs and select the models with the highest R@1 value on the test set, respectively. From the results, the model trained on the randomly sampled dataset gets the worst performance than those trained on our YouTube dataset. The results demonstrate that our neat dataset indeed can facilitate the learning of basic game concepts. A notable observation is that [ours] achieves higher success rates on RL tasks but lower performance on retrieval tasks compared to [w/o LCF], suggesting that the video-text alignment objective does not fully align with RL requirements. [w/o LCF] is trained on the dataset filtered only by global-level correlation, which is directly conducted at the video level. In contrast, [ours] is trained on the dataset further filtered by local-level correlation, which is object-centric designing for RL training. \textit{Note that [w/o LCF] is the dataset we plan to release for better training the VLM model for video-text retrieval tasks on Minecraft.} 

\section{Conclusion}
We construct a neat YouTube dataset based on the large-scale YouTube database provided by MineDojo. Moreover, we introduce a novel cross-modal contrastive learning framework architecture, CLIP4MC, to learn an RL-friendly VLM that serves as an intrinsic reward function for open-ended tasks. Our findings suggest that our dataset enhances the acquisition of fundamental game concepts and CLIP4MC delivers a more effective reward signal for RL training.

\section*{Acknowledgements}
This work was supported by NSFC under grant 62250068. The authors would like to thank the anonymous reviewers for their valuable comments.

\bibliographystyle{splncs04}
\bibliography{main}

\appendix
\onecolumn

\section{Dataset Details}
\label{app:dataset}

\subsection{Key Entity Size Estimation}

MaskCLIP \cite{zhou2022extract} has shown notable zero-shot segmentation ability of CLIP \cite{RadfordKHRGASAM21} without fine-tuning. Given that the architecture and training loss of MineCLIP \cite{fan2022minedojo} are similar to those of CLIP, we hypothesize that MineCLIP could also achieve zero-shot segmentation. Therefore, we modify the vision transformer in MineCLIP following MaskCLIP, replacing the output of \texttt{CLS} token with value-embeddings of all patches in the final self-attention block. This change enables the vision transformer to output embedding vectors for each image patch rather than a single vector for the entire image. Since there is an additional temporal transformer in MineCLIP visual encoder compared to CLIP, we pass the patch embedding vectors through this temporal transformer individually, ensuring the patch embeddings are aligned with MineCLIP's embedding space. With an image input, we now acquire embedding vectors for each patch in MineCLIP's embedding space. We also generate text embedding vectors for all entities in the MineDojo keyword list using the MineCLIP text encoder. The cosine similarities between each patch embedding vector and the text embedding vectors are then calculated. 

\begin{figure}[h]
    \centering
    \begin{subfigure}{0.23\textwidth}
        \centering
        \includegraphics[width=\textwidth]{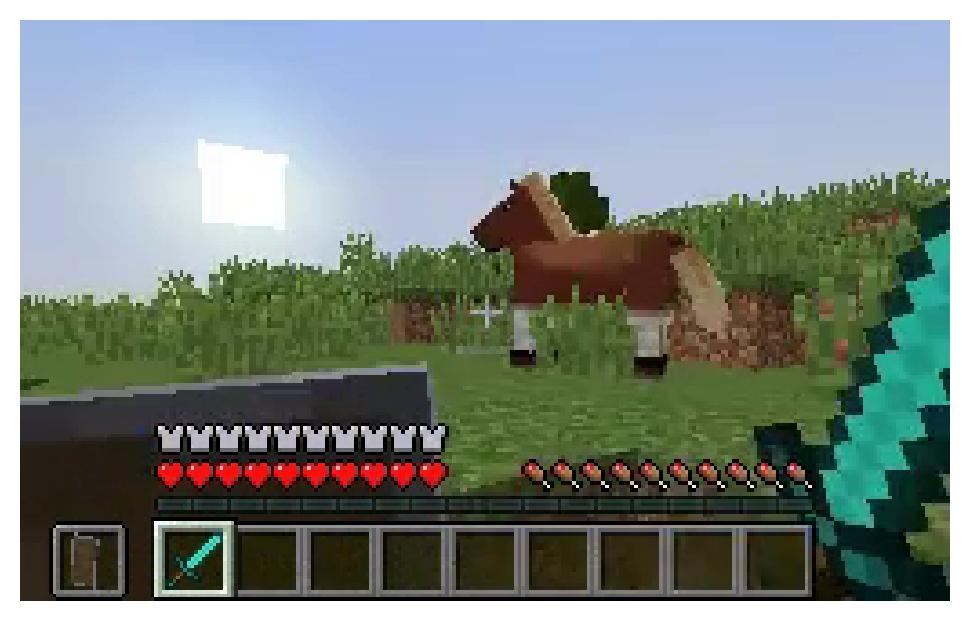}
        \caption{RGB image}
        \label{fig:bbox_eg_rgb}
    \end{subfigure}
    \hfill
    \begin{subfigure}{0.23\textwidth}
        \centering
        \includegraphics[width=\textwidth]{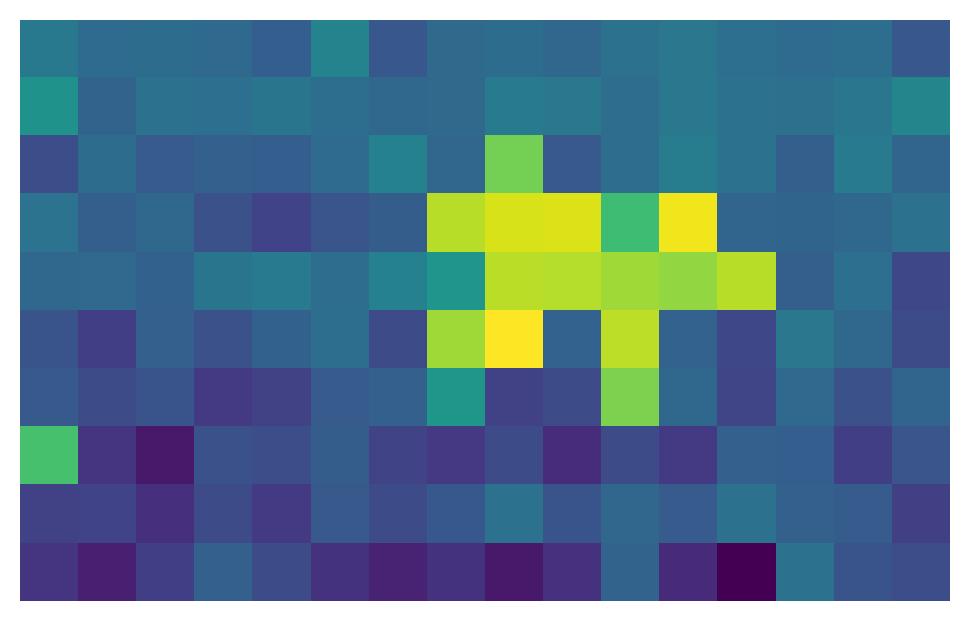}
        \caption{Similarity scores}
        \label{fig:bbox_eg_logit}
    \end{subfigure}
    \hfill
    \begin{subfigure}{0.23\textwidth}
        \centering
        \includegraphics[width=\textwidth]{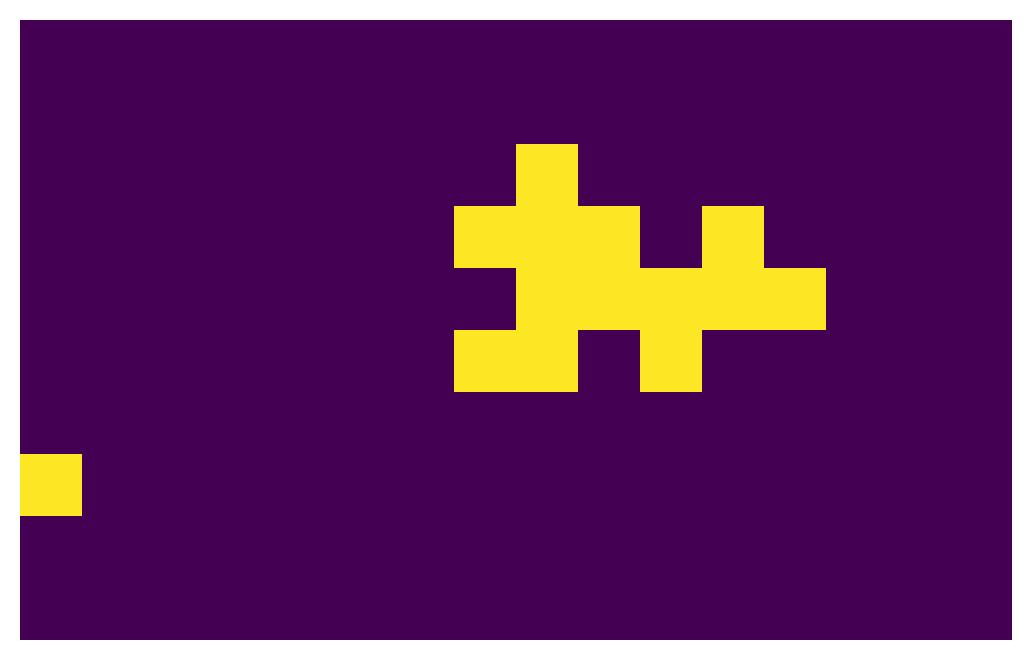}
        \caption{Filtering}
        \label{fig:bbox_eg_prob_bi}
    \end{subfigure}
    \hfill
    \begin{subfigure}{0.23\textwidth}
        \centering
        \includegraphics[width=\textwidth]{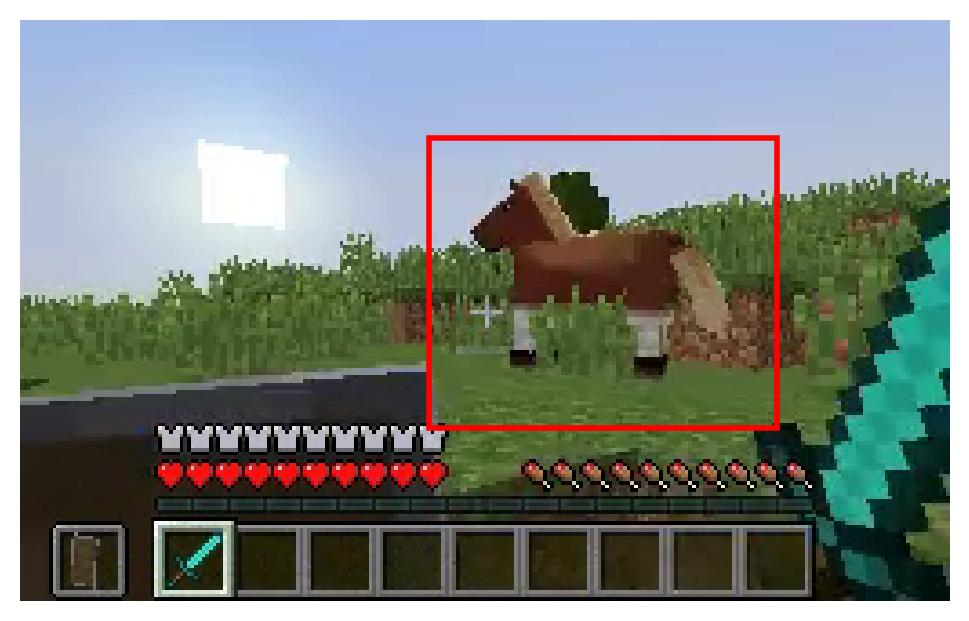}
        \caption{Bounding box}
        \label{fig:bbox_eg_bbox}
    \end{subfigure}
    \caption{Procedure of generating a bounding box for the key entity. (a) The raw RGB image containing the key entity ``horse''. (b) The heatmap illustrating the similarity scores of ``horse'' on each patch. (c) The filtered heatmap. (d) The bounding box of ``horse''.}
    \label{fig:bbox_eg}
    \vspace{-4mm}
\end{figure}

As our goal is to estimate the size of the key entity mentioned in the corresponding transcript, we extract the similarity scores of the key entity on each image patch and create a heatmap for it. For instance, \Cref{fig:bbox_eg_logit} shows a heatmap generated for the entity ``horse''. To refine this heatmap, we apply two filtering criteria: firstly, a patch is retained only if the key entity's similarity score on it is higher than that of any other entity in the keyword list; secondly, we ignore patches with similarity scores below a threshold, $\tau_{\rm patch}=0.295$, to reduce noise. After filtering, we find the maximum connected region in the filtered heatmap and compute the minimum bounding box that covers this region, as depicted in \Cref{fig:bbox_eg_bbox}. This bounding box allows us to estimate the size of the key entity within the image, calculated as the area of the bounding box $H_bW_b$.

\subsection{Construction Process}

The construction steps of the dataset are as follows:

\begin{enumerate}
\setlength\itemsep{1mm}
    \item Obtain YouTube videos along with their transcripts and the keywords list from the MineDojo database\footnote{\url{https://github.com/MineDojo/MineDojo}}.
    \item Use Transcript Cleaning (\Cref{sec:tc}) to enhance the accuracy of transcripts and ensure they are complete sentences.
    \item For each video with a transcript, annotate all keywords (including different forms of keywords such as combined words, plural forms, etc.) appearing in the transcript.
    \item Use Keyword Filtering (\Cref{sec:kf}) to find all transcript sentences containing keywords. The length of sentences is adjusted to 10-35 words.
    \item Extract all non-overlapping transcript clips from each video with a transcript following step 4.
    \item For each transcript clip, calculate the central timestamp corresponding to the clip based on the transcript timestamps. Use this central timestamp to extract a video clip of duration $D$ seconds from the video.
    \item From all the video-clip pairs extracted in the previous steps, extract $M$ pairs. Utilize Video Partition and Selection (\Cref{sec:vps}) to select one from $p$ partitions, in order to handle the issue of scene transitions and thereby mitigate interference from other extraneous information.
    \item Use Correlation Filtering (\Cref{sec:ccf}) to select the top $k\%$ pairs with the highest correlation from the $M$ pairs as the training set. Specifically, we firstly select pairs following the local correlation filtering criteria, \ie, the size of target entity. If the number of video clips with explicit target entities is less than $k\%$, we select the remaining training pairs based on the global correlation filtering criteria, \ie, the cosine similarity calculated via MineCLIP.
    \item Randomly select $M^\prime$ pairs in addition to the $M$ pairs as the test set.
\end{enumerate}

The parameters used in the above process are listed in Table~\ref{tab:parameters}.
Following this process, we construct a training set of size 640K and a test set of size 4096. This is the dataset for better training the VLM for RL tasks on Minecraft. Our other dataset for better training the VLM for video-text retrieval tasks on Minecraft is only filtered by the global correlation filtering criteria to select the top $k\%$ pairs in step 8. We have released the two datasets by specifying the transcript clips and the corresponding timestamps of the videos in the original database. The link is \hyperlink{https://huggingface.co/datasets/AnonymousUserCLIP4MC/CLIP4MC}{https://huggingface.co/datasets/AnonymousUserCLIP4MC/CLIP4MC}.

\begin{table}[h]\centering 
\caption{Values of parameters used in the dataset construction process.}
\renewcommand{\arraystretch}{1.2}
\begin{footnotesize}
\begin{tabular}{@{}cc@{}} 
\toprule
\textbf{Parameter} & \textbf{Value}\\ \midrule
$D$ & 16 \\
$M$ & 1,280,000  \\
$p$ & 3 \\
$k$ & 50    \\
$M^\prime$ & 4,096\\ \bottomrule
\end{tabular}
\end{footnotesize}
\label{tab:parameters}
\end{table}

\begin{table}[htbp]\centering 
\caption{Training hyperparameters for CLIP4MC.}
\renewcommand{\arraystretch}{1.2}
\begin{tabular}{@{}cc@{}} 
\toprule
\textbf{Hyperparameter} & \textbf{Value}\\ \midrule
warmup steps & 320 \\
learning rate & 1.5e-4 \\
lr schedule & cosine with warmup \\
weight decay & 0.2 \\
layerwise lr decay & 0.65 \\
batch size per GPU & 100 \\
parallel GPUs & 4 \\
video resolution & 160 × 256 \\
number of frames & 16 \\
image encoder & ViT-B/16 \\
swap threshold $\tau$ & 0.02 \\
\bottomrule
\end{tabular}
\label{tab:hyperparameters}
\vspace{-2mm}
\end{table}

\section{Architecture}
\label{app:arch}

\textbf{Input Details.}
The length of each transcript clip is 25 words, while the length of the video is 16 seconds. The resolution of the video stream is 160 × 256, with 5 fps. In other words, the video stream is 80 frames. As for the video snippet, we further equidistantly sample it to 16 frames for fewer computing resources.

\vspace{1mm}
\noindent \textbf{Text Encoder Details.}
Referring to the design of MineCLIP \cite{fan2022minedojo}, the text encoder is a 12-layer 512-width GPT model, which has 8 attention heads. The textual input is tokenized via the tokenizer used in CLIP and is padded/truncated to 77 tokens. The initial weights of the model use the public checkpoint of CLIP and only finetune the last two layers during training.

\vspace{1mm}
\noindent \textbf{Spatial Encoder Details.}
The Spatial encoder is a frame-wise image encoder referred to the design of MineCLIP \cite{fan2022minedojo}, which uses the ViT-B/16 architecture to compute a 512-D embedding for each frame. The initial weights of the model use the public checkpoint of OpenAI CLIP, and only the last two layers are finetuned during training.

\vspace{1mm}
\noindent \textbf{Temporal Transformer Details.}
The temporal Transformer is used to aggregate the temporal information from the spatial encoder, which is a 2-depth 8-head Transformer module whose input dimension is 512 and maximum sequence length is 32.

\vspace{1mm}
\noindent \textbf{Adapter Details.}
In order to get better embedding, we use an adapter to map video embedding and text embedding. The adapter models are 2-layer MLP, except the text adapter which is an identity. 

\section{CLIP4MC Training}
\label{app:train}

The training process for CLIP4MC is adapted from the training processes for CLIP4Clip \cite{LuoJZCLDL22} and MineCLIP \cite{fan2022minedojo}. We trained all models on the 640K training set. For each video-text clip pair, we obtain 16 frames of RGB image through equidistant sampling and normalize each channel separately. During training, we use random resize crops for data augmentation. We use cosine learning rate annealing with 320 gradient steps of warming up. We only fine-tune the last two layers of pre-trained CLIP encoders, and we apply a module-wise learning rate decay (learning rate decays along with the modules) for better fine-tuning. Training is performed on 1 node of 4 × A100 GPUs with FP16 mixed precision via the PyTorch native \texttt{amp} module. All hyperparameters are listed in Table~\ref{tab:hyperparameters}.

\section{Environment Details}
\label{app:env}

\subsection{Environment Initialization}

\Cref{tab:setup} outlines how we set up and initialize the environment for each task in our RL experiments. \textbf{SR} denotes the spawn range of animals, and \textbf{Length} represents the length of one episode. All tasks are in the biome \texttt{plains}.

\begin{table}[h]
    \centering 
    \caption{Environment setup in our experiments.}
    \label{tab:setup}
    \renewcommand{\arraystretch}{1.2}
    \begin{tabular}{lcccc p{0.35\columnwidth}} 
    \toprule
    \multicolumn{1}{c}{\textbf{Task}} & \textbf{Mob} & \textbf{SR} & \textbf{Length} & \textbf{Tool} & \multicolumn{1}{c}{\textbf{Prompt}} \\ 
    \midrule
    milk a cow & cow & 10 & 200 & empty\_bucket & \textit{obtain milk from a cow in plains with a bucket} \\
    shear wool & sheep & 10 & 200 & shears & \textit{shear a sheep in plains with shears} \\
    combat a spider & spider & 7 & 500 & diamond\_sword & \textit{combat a spider in plains with a diamond sword} \\
    combat a zombie & zombie & 7 & 500 & diamond\_sword & \textit{combat a zombie in plains with a diamond sword} \\
    hunt a cow & cow & 7 & 500 & diamond\_sword & \textit{hunt a cow in plains with a diamond sword} \\
    hunt a sheep & sheep & 7 & 500 & diamond\_sword & \textit{hunt a sheep in plains with a diamond sword} \\
    hunt a pig & pig & 7 & 500 & diamond\_sword & \textit{hunt a pig in plains with a diamond sword} \\
    hunt a chicken & chicken & 7 & 500 & diamond\_sword & \textit{hunt a chicken in plains with a diamond sword} \\
    \bottomrule
    \end{tabular}
\end{table}

\subsection{Observation Space}

To enable the creation of multi-task and continually learning agents that can adapt to new scenarios and tasks, MineDojo provides unified observation and action spaces \cite{fan2022minedojo}. \Cref{tab:obs} provides detailed descriptions of the observation space adopted in our experiments.

\begin{table}[h!]
\centering
\caption{Observation space in our experiments.}
\label{tab:obs}
\begin{small}
\setlength{\tabcolsep}{5pt}
\begin{tabular}{l p{0.6\textwidth}}
    \toprule
    \textbf{Observation} & \textbf{Descriptions} \\
    \midrule
    \textbf{Egocentric RGB frame} &
      \begin{tabular}[c]{@{}p{0.6\textwidth}@{}}RGB frames provide an egocentric view of the running Minecraft client; The shape height (H) and width (W) are specified by argument image\_size. In our experiment, it is (160, 256).\end{tabular} \\
     &
       \\
    \textbf{Voxels} &
      \begin{tabular}[c]{@{}p{0.6\textwidth}@{}}Voxels observation refers to the 3x3x3 surrounding blocks around the agent. This type of observation is similar to how human players perceive their surrounding blocks. It includes names and properties of blocks.\end{tabular} \\
       \\
    \textbf{Location Statistics} &
      \begin{tabular}[c]{@{}p{0.6\textwidth}@{}}Location statistics include information about the terrain the agent currently occupies. It also includes the agent’s location and compass.\end{tabular} \\
     &
       \\
    \textbf{Biome ID} & \begin{tabular}[c]{@{}p{0.6\textwidth}@{}}Index of the environmental biome where the agent spawns.\end{tabular} \\
    \bottomrule
\end{tabular} 
\vspace{-2mm}
\end{small}
\end{table}

\subsection{Action Space}

We simplify the original action space of MineDojo \cite{fan2022minedojo} into a two-dimensional multi-discrete action space. The first dimension consists of 12 discrete actions: NO\_OP, forward, backward, left, right, jump, sneak, sprint, camera pitch +30, camera pitch -30, camera yaw +30, and camera yaw -30. The second dimension contains NO\_OP, attack, and use.

\section{RL Training}
\label{app:rl}

\subsection{Agent Network}

Like MineAgent \cite{fan2022minedojo}, our policy framework is also composed of three components: an encoder for input features, a policy head, and a value head. In order to deal with cross-modal observations, the feature extractor includes a variety of modality-specific components:

\vspace{2mm} \noindent \textbf{RGB frame.} To optimize for computational efficiency and equip the agent with strong visual representations from scratch, we use the fixed frame-wise image encoder $\psi _I$ from CLIP4MC to process RGB frames.

\vspace{2mm} \noindent \textbf{Goal of the task.} The text embedding of the natural language task objective (prompt) is computed by $\psi _G$ from CLIP4MC.

\vspace{2mm} \noindent \textbf{Yaw and Pitch.} We first compute sin(·) and cos(·) features respectively, then pass them through CompassMLP as described in \Cref{tab:network}.

\vspace{2mm} \noindent \textbf{GPS.} The position coordinates are normalized and passed through GPSMLP as described in \Cref{tab:network}.

\vspace{2mm} \noindent \textbf{Voxels.} To process the 3 × 3 × 3 voxels surrounding the agent, we embed discrete block names as dense vectors, flatten them, and pass them through VoxelEncoder as described in \Cref{tab:network}.

\vspace{2mm} \noindent \textbf{Previous action.} Our agent relies on its immediate previous action, which is embedded and processed through Prev Action Emb as described in \Cref{tab:network}, which is a conditioning factor.

\vspace{2mm} \noindent \textbf{BiomeID.} To perceive the discrepancy in different environments, we embed BiomeID as a vector through an MLP named BiomeIDEmb as described in \Cref{tab:network}. 

\vspace{2mm}

The features from each modality are combined by concatenation, passed through an additional feature fusion network as described in \Cref{tab:network}, and then fed into a GRU to integrate historical information. The output of GRU serves as the input for both the policy head and the value head. The policy head is modeled using an MLP, as described in \Cref{tab:network} for Actor \& Critic Network, which transforms the input feature vectors into an action probability distribution. Similarly, value MLP is used to estimate the value function, conditioned on the same input features.

\begin{table}[h]
\centering
\caption{Policy Network Details}
\label{tab:network}
\renewcommand{\arraystretch}{1.3}
\setlength{\tabcolsep}{5pt}
\begin{small}
\begin{tabular}{cp{0.7\textwidth}}
\toprule
\textbf{Network} & \textbf{Details} \\ \midrule
\textbf{Actor \& Critic Net} & hidden\_dim: 256, hidden\_depth: 2 \\
\textbf{GRU} & hidden\_dim: 256, layer\_num: 1 \\
\textbf{CompassMLP} & hidden\_dim: 128, output\_dim: 128, hidden\_depth: 2 \\
\textbf{GPSMLP} & hidden\_dim: 128, output\_dim: 128, hidden\_depth: 2 \\
\textbf{VoxelEncoder} & embed\_dim: 8, hidden\_dim: 128, output\_dim: 128, hidden\_depth: 2 \\
\textbf{BiomeIDEmb} & embed\_dim: 8 \\
\textbf{PrevActionEmb} & embed\_dim: 8 \\
\textbf{ImageEmb} & output\_dim: 512 \\
\textbf{FeatureFusion} & output\_dim: 512, hidden\_depth: 0 \\ 
\bottomrule
\end{tabular}
\end{small}
\end{table}

\subsection{Algorithm}

In our experiments, we implement Proximal Policy Optimization (PPO) \cite{schulman2017proximal} as our base RL algorithm. In detail, PPO updates policies via
\begin{equation*}
    \theta_{k+1} = \arg\max_{\theta} \mathbb{E}_{s,a\sim \pi_{\theta_k}}\left[L(s,a,\theta_k,\theta)\right].
\end{equation*}
Here, $L$ is defined as
\begin{equation*}
    L(s,a,\theta_k,\theta) = \min\left(\frac{\pi_\theta(a|s)}{\pi_{\theta_k}(a|s)}A^{\pi_{\theta_k}}(s,a),\operatorname{clip}\left(\frac{\pi_\theta(a|s)}{\pi_{\theta_k}(a|s)}, 1-\epsilon, 1+\epsilon\right)A^{\pi_{\theta_k}}(s,a)\right),
\end{equation*}
where $\epsilon$ is a clip value that limits the deviation between the new policy $\pi_{\theta_{k+1}}$ and the old $\pi_{\theta_k}$. $A^{\pi_{\theta_k}}(s,a)$ is the advantage function for the current policy and estimated via Generalized Advantage Estimation (GAE) \cite{schulman2015high}. Unlike MineDojo \cite{fan2022minedojo}, our implementation does not include additional self-imitation learning and action smoothing techniques as we found that our adopted PPO is already able to achieve high performance. The hyperparameters of our PPO implementation are listed in \Cref{tab:ppo} and shared across all experiments.

\begin{table}[h]
    \centering
    \caption{PPO hyperparameters in our experiments.}
    \label{tab:ppo}
    \begin{footnotesize}
    \renewcommand{\arraystretch}{1.2}
    \begin{tabular}{cc}
        \toprule
        \textbf{Hyperparameter} & \textbf{Value} \\
        \midrule
        num steps & 1000 \\
        num envs & 4 \\
        num minibatches & 4 \\
        num epochs & 8 \\
        GAE lambda & 0.95 \\
        discounted gamma & 0.99 \\
        entropy coef & 0.005 \\
        clip epsilon & 0.02 \\
        learning rate & 1e-4 \\
        optimizer & Adam \\
        GRU data chunk length & 10 \\
        gradient norm & 10.0 \\
        \bottomrule
    \end{tabular}
    \end{footnotesize}
\end{table}

\subsection{Coefficient of MineCLIP Reward}

\begin{figure}[t]
    \centering
    \begin{subfigure}{0.3\textwidth}
        \centering
        \includegraphics[width=\textwidth]{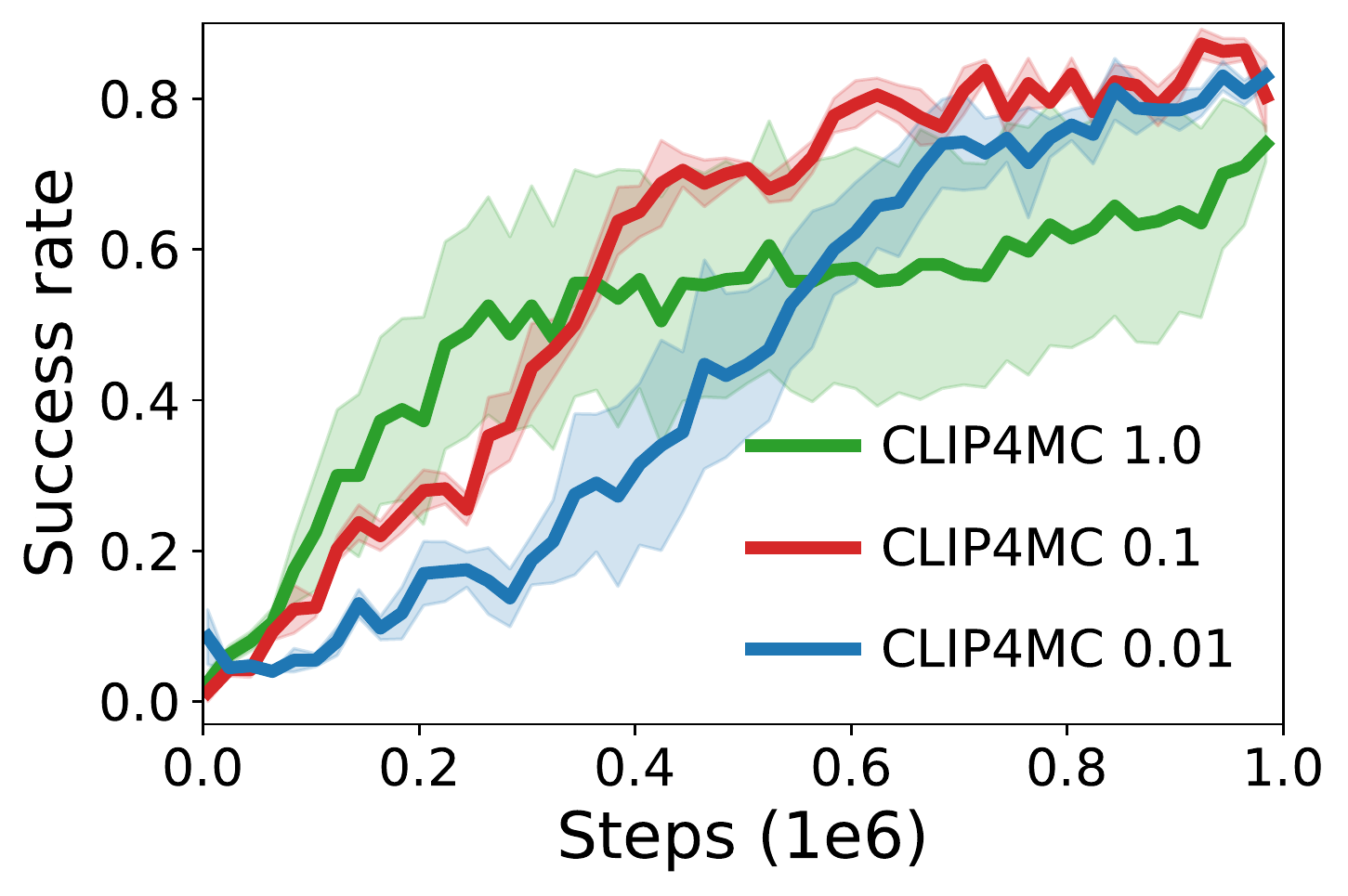}
        \caption{CLIP4MC}
    \end{subfigure}
    \begin{subfigure}{0.3\textwidth}
        \centering
        \includegraphics[width=\textwidth]{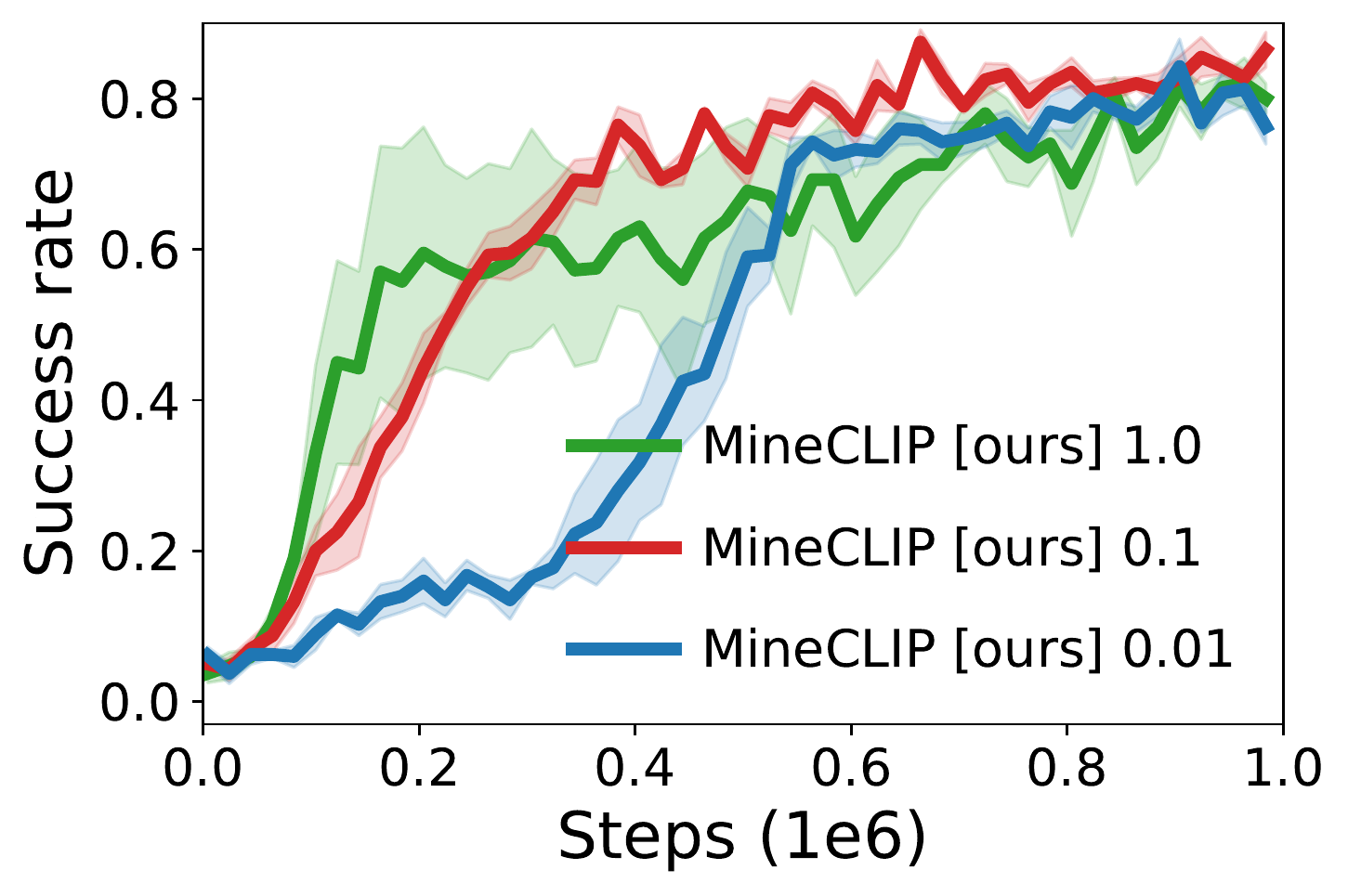}
        \caption{MineCLIP [ours]}
    \end{subfigure}
    \begin{subfigure}{0.3\textwidth}
        \centering
        \includegraphics[width=\textwidth]{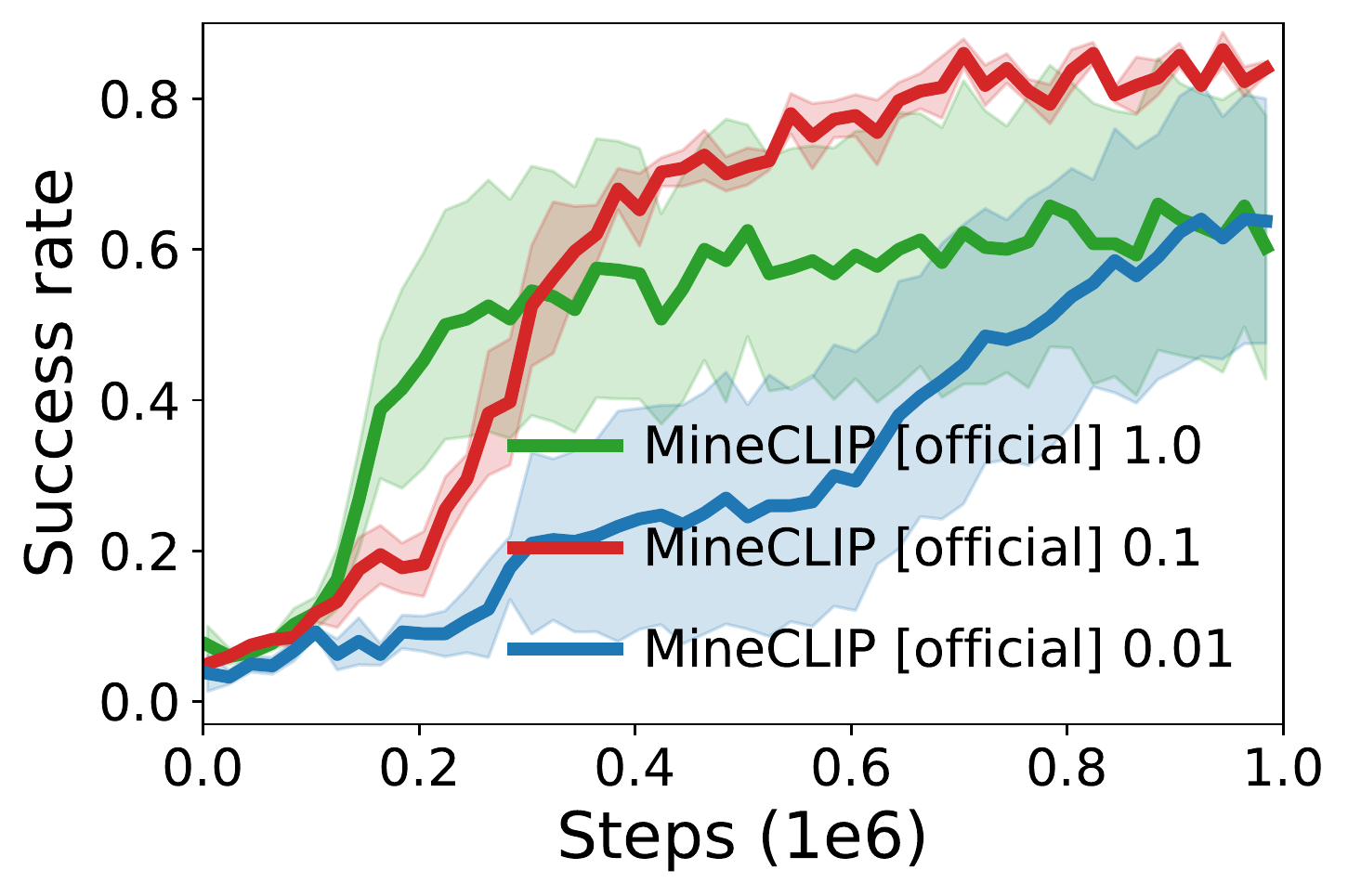}
        \caption{MineCLIP [official]}
    \end{subfigure}
    \caption{Learning curves of CLIP4MC, MineCLIP [ours], and MineCLIP [official] with different reward coefficients on task \textit{milk a cow}.}
    \label{fig:milk_coef}
\end{figure}

\begin{figure}[t]
    \centering
    \begin{subfigure}{0.3\textwidth}
        \centering
        \includegraphics[width=\textwidth]{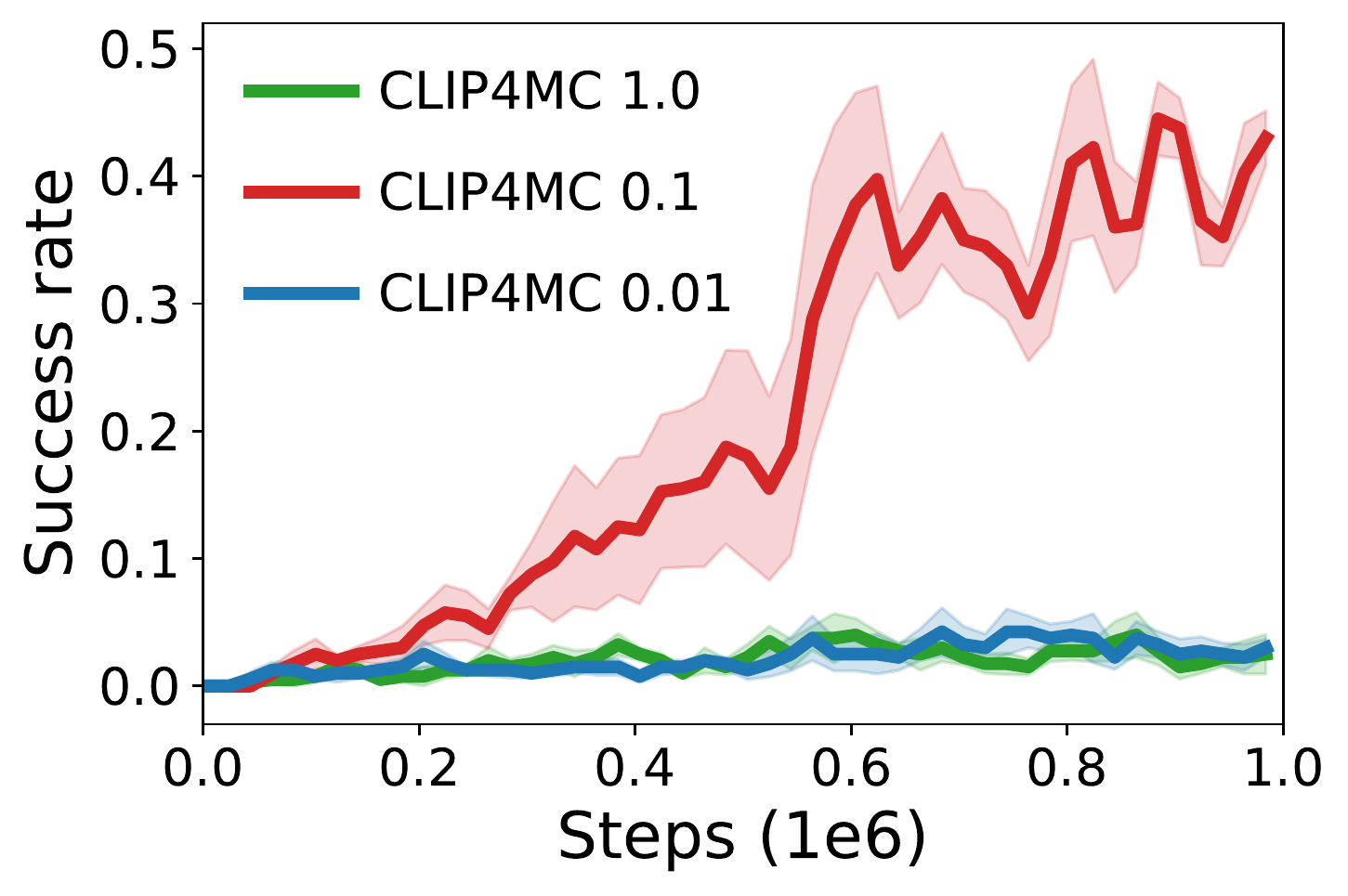}
        \caption{CLIP4MC}
    \end{subfigure}
    \begin{subfigure}{0.3\textwidth}
        \centering
        \includegraphics[width=\textwidth]{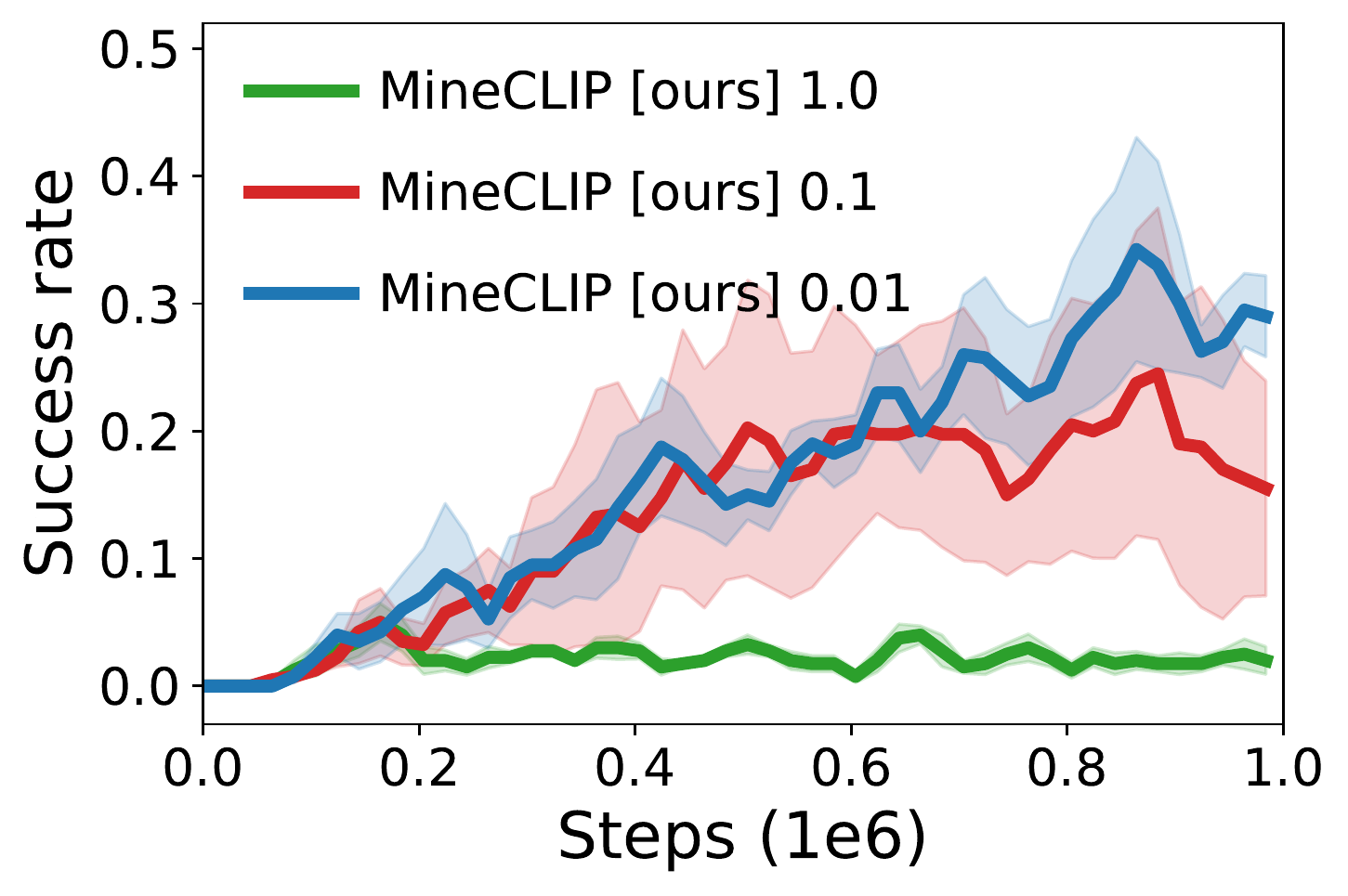}
        \caption{MineCLIP [ours]}
    \end{subfigure}
    \begin{subfigure}{0.3\textwidth}
        \centering
        \includegraphics[width=\textwidth]{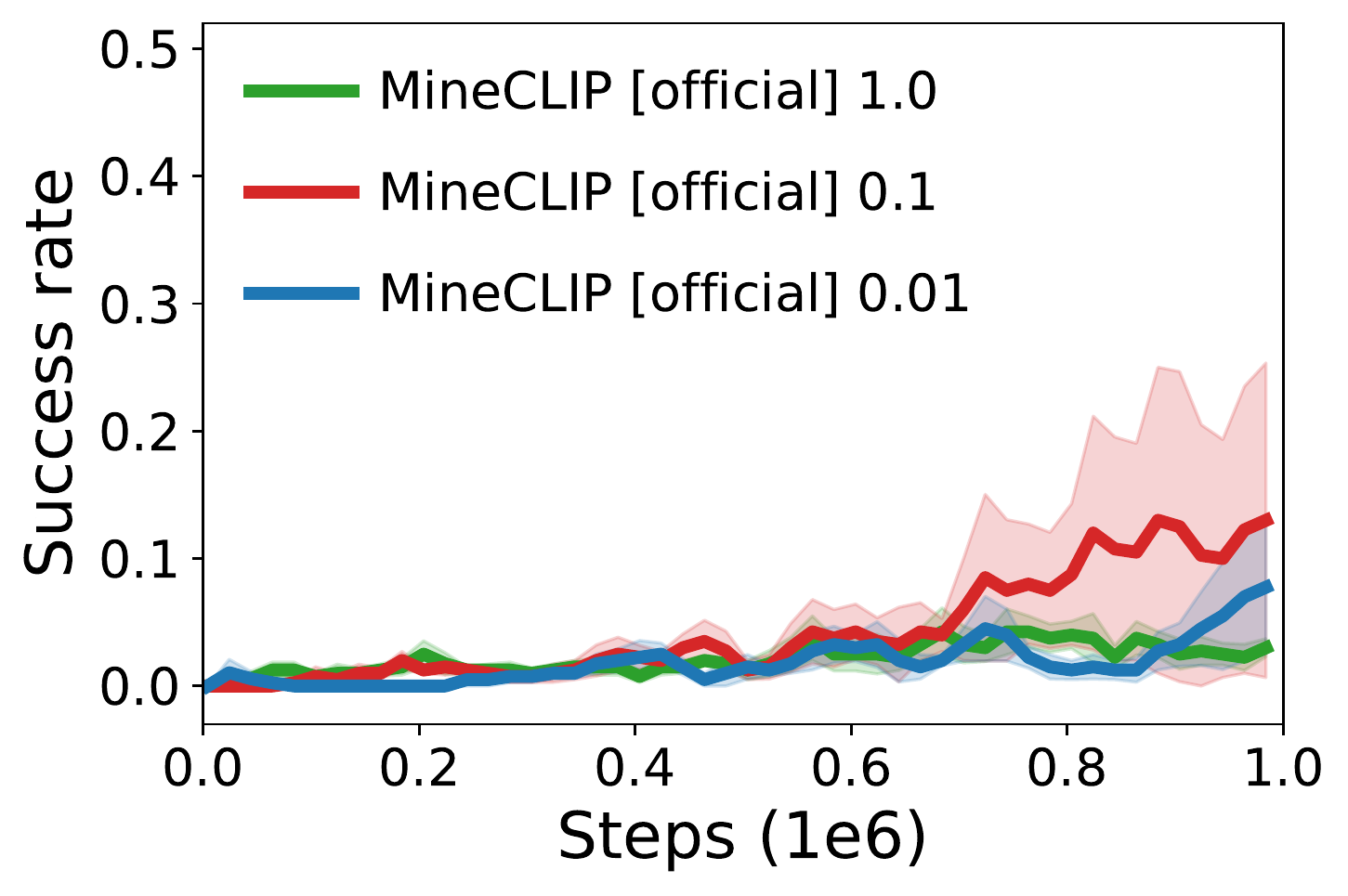}
        \caption{MineCLIP [official]}
    \end{subfigure}
    \caption{Learning curves of CLIP4MC, MineCLIP [ours], and MineCLIP [official] with different reward coefficients on task \textit{hunt a cow}.}
    \label{fig:cow_coef}
\end{figure}

In our experiments, the agent is trained with the shaped reward defined as $r_t = r_t^{\rm env} + c \cdot r_t^{\rm mc}$. Specifically, $r_t^{\rm env}$ represents the sparse reward provided by the environment. The environment will provide a +100 reward only when the agent successfully accomplishes the task; otherwise, the environment reward is zero. The other component, $r_t^{\rm mc}$, denotes the MineCLIP reward introduced in \Cref{sec:reward}. The coefficient $c$ controls the weight of the MineCLIP reward. For RL algorithms in sparse-reward environments, the coefficient of the intrinsic reward has a significant influence on the final performance of algorithms. Therefore, we conduct preliminary experiments to decide a suitable coefficient of MineCLIP reward for our experiments in the paper. We choose two tasks, \textit{milk a cow} and \textit{hunt a cow}, as our testbed. As shown in \Cref{fig:milk_coef,fig:cow_coef}, three values, $c=1.0$, $c=0.1$, and $c=0.01$, are evaluated on CLIP4MC, MineCLIP [ours], and MineCLIP [official]. The results demonstrate that $c=0.1$ achieves the best performance in 5 out of 6 experiments. Based on this observation, we set $c=0.1$ for all experiments in this paper.

\subsection{Learning Curves}

\begin{figure}[t]
    \centering 
    \begin{subfigure}{0.25\textwidth}
		\centering
		\includegraphics[width=\textwidth]{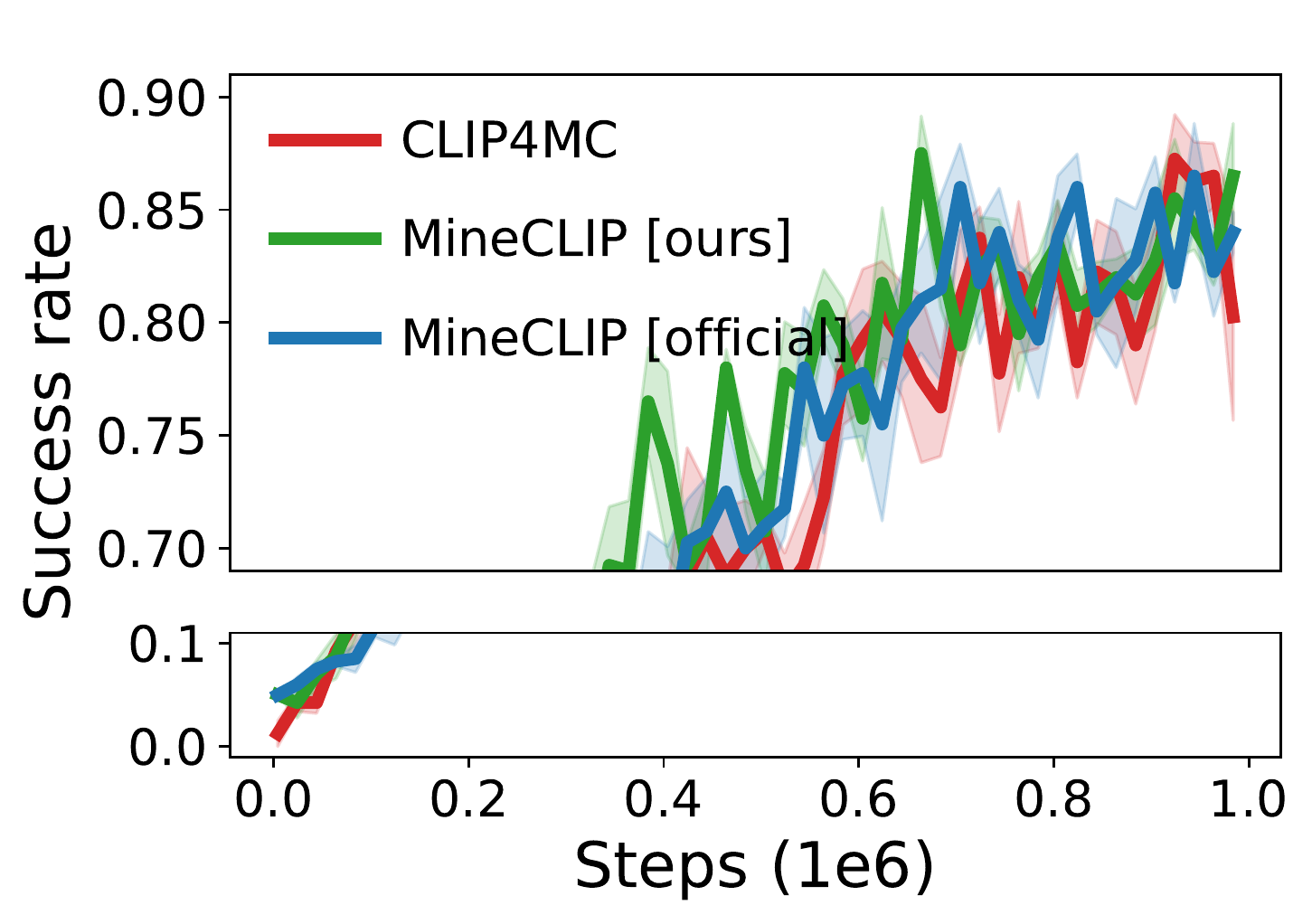}
		\caption{milk a cow}
        \vspace{2mm}
    \end{subfigure}
    \hspace{-0.1in}
    \begin{subfigure}{0.25\textwidth}
		\centering
		\includegraphics[width=\textwidth]{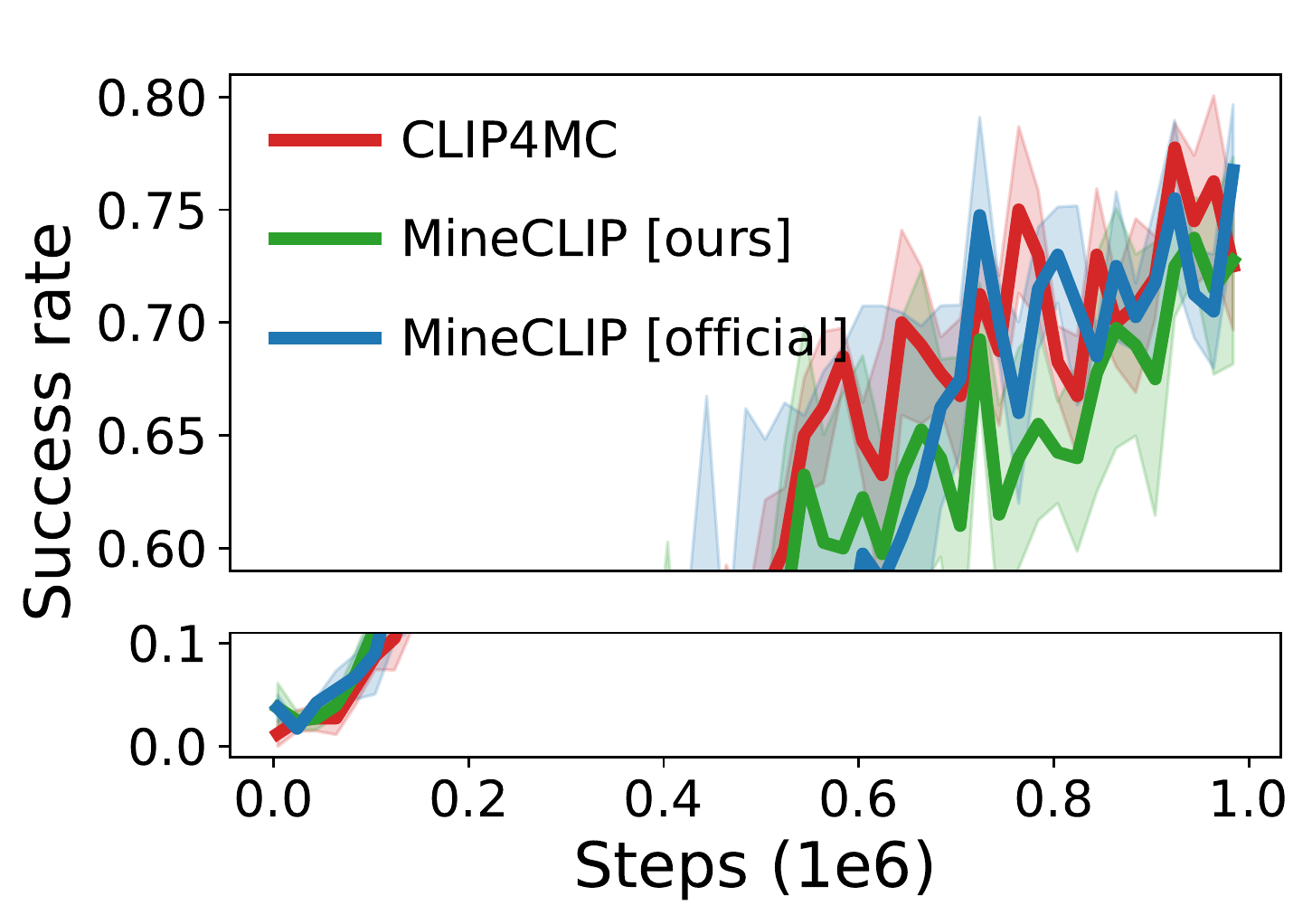}
		\caption{shear wool}
        \vspace{2mm}
    \end{subfigure}
    \hspace{-0.1in}
    \begin{subfigure}{0.25\textwidth}
		\centering
		\includegraphics[width=\textwidth]{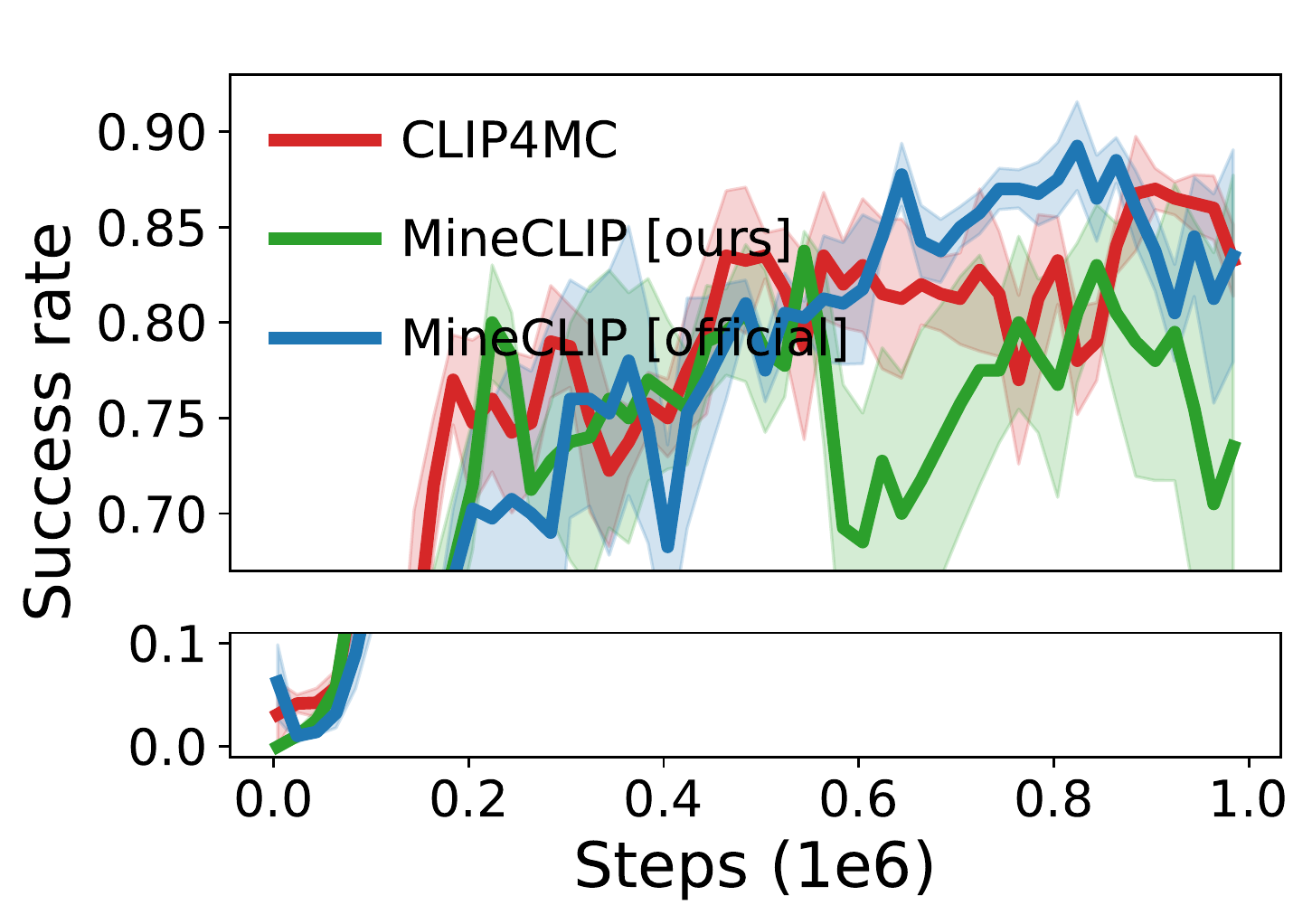}
		\caption{combat a spider}
        \vspace{2mm}
    \end{subfigure}
    \hspace{-0.1in}
    \begin{subfigure}{0.25\textwidth}
		\centering
		\includegraphics[width=\textwidth]{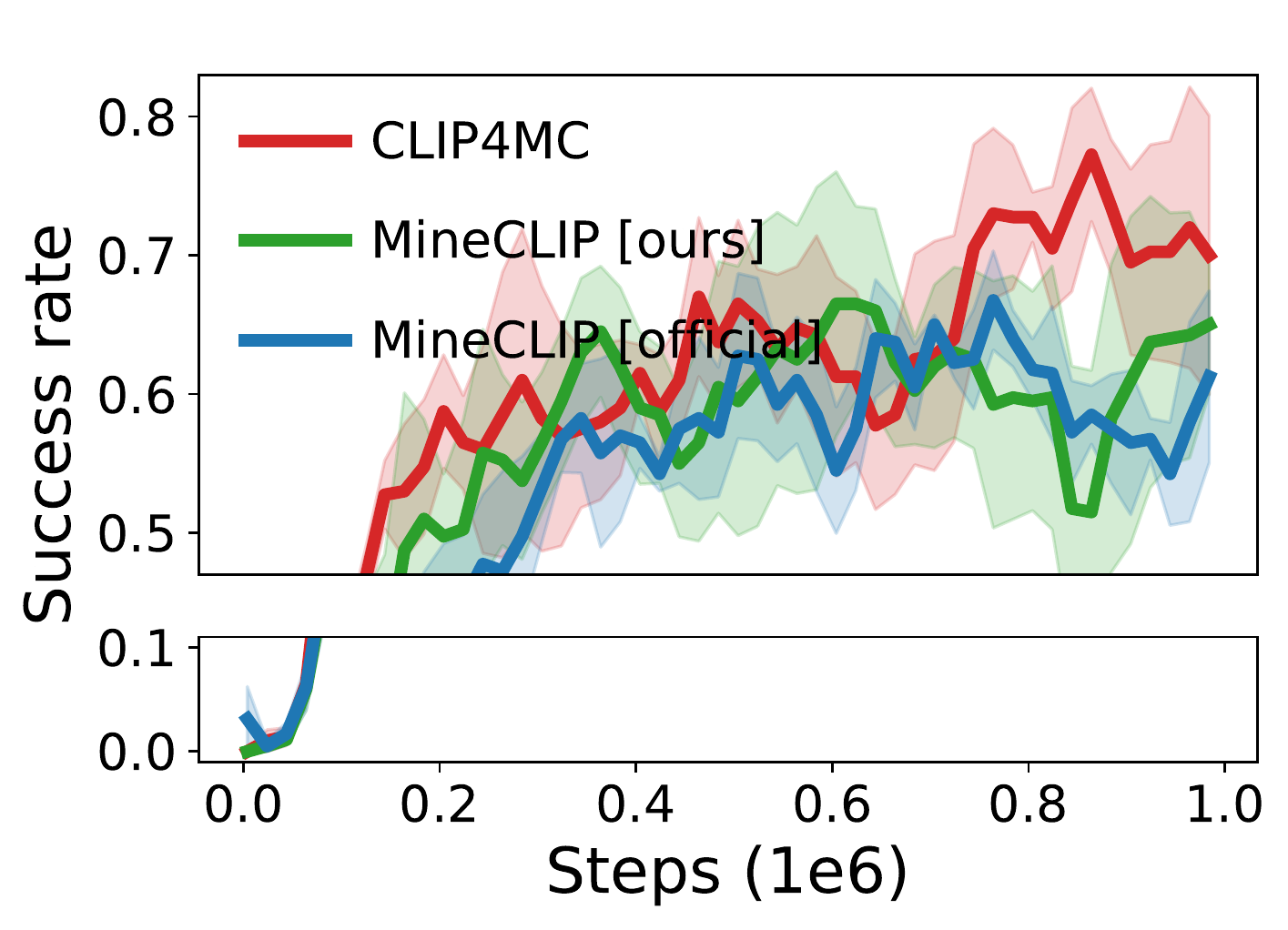}
		\caption{combat a zombie}
        \vspace{0.08in}
    \end{subfigure}
    \begin{subfigure}{0.25\textwidth}
		\centering
		\includegraphics[width=\textwidth]{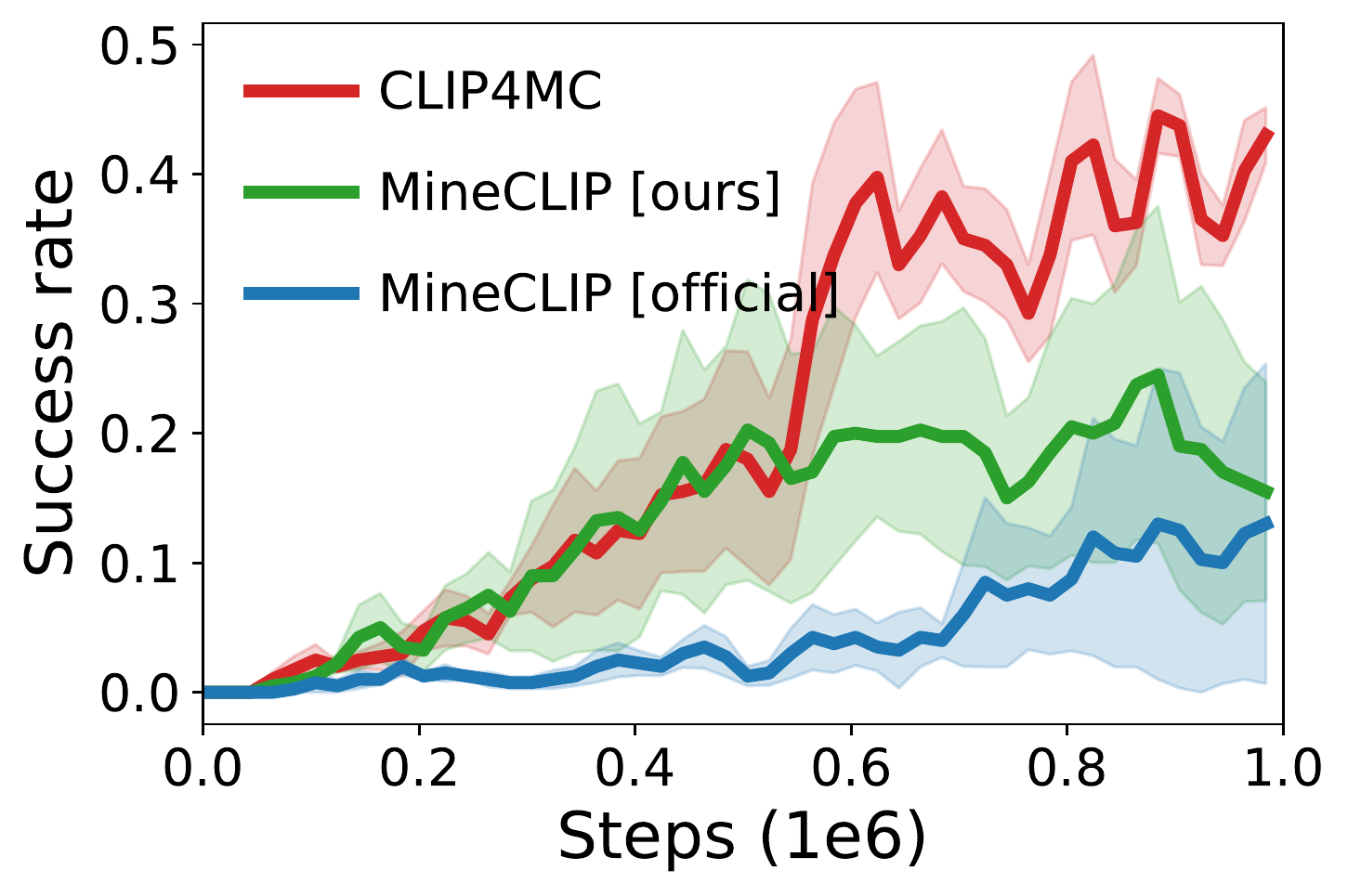}
		\caption{hunt a cow}
    \end{subfigure}
    \hspace{-0.1in}
    \begin{subfigure}{0.25\textwidth}
		\centering
		\includegraphics[width=\textwidth]{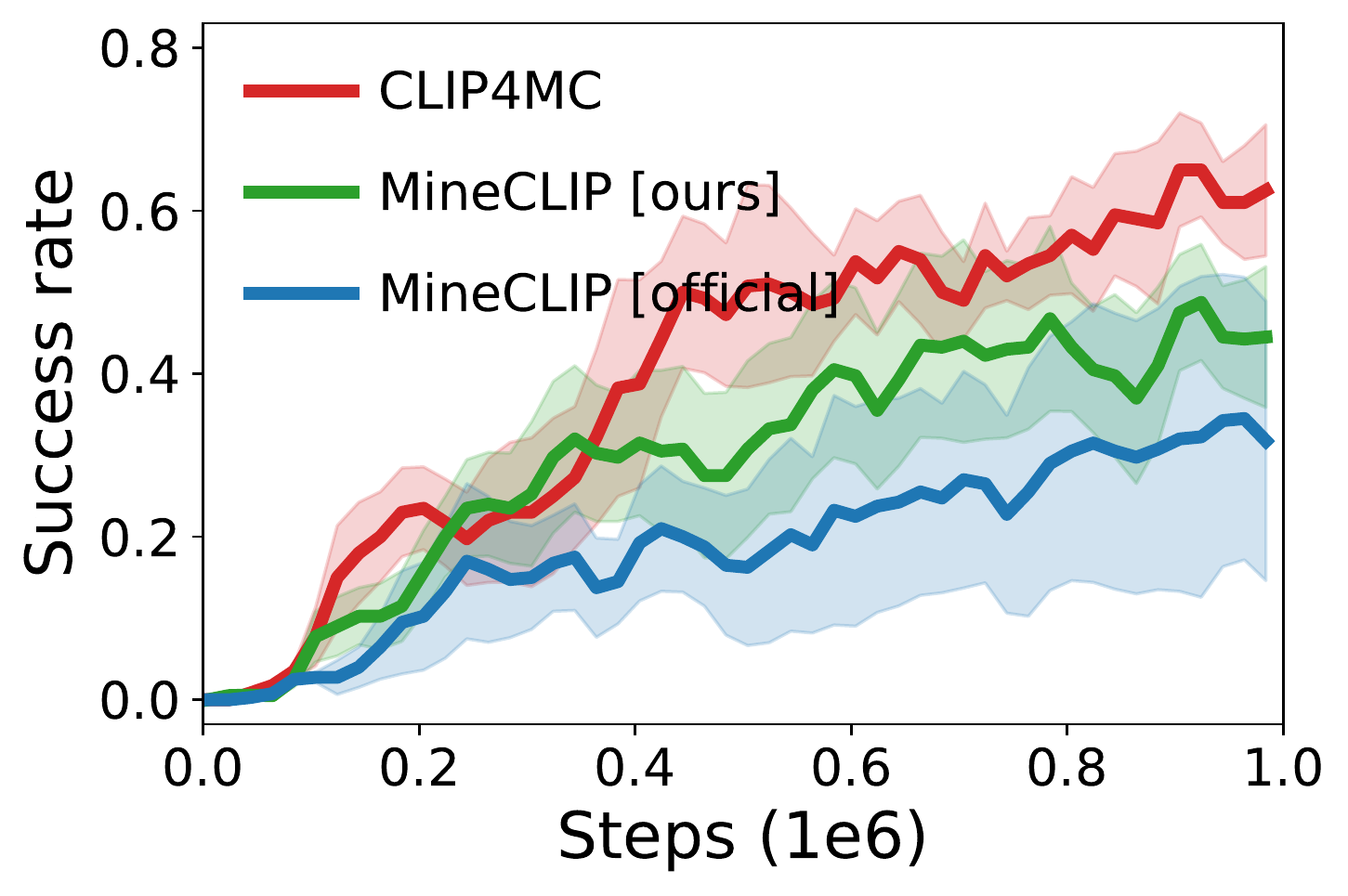}
		\caption{hunt a sheep}
    \end{subfigure}
    \hspace{-0.1in}
    \begin{subfigure}{0.25\textwidth}
		\centering
		\includegraphics[width=\textwidth]{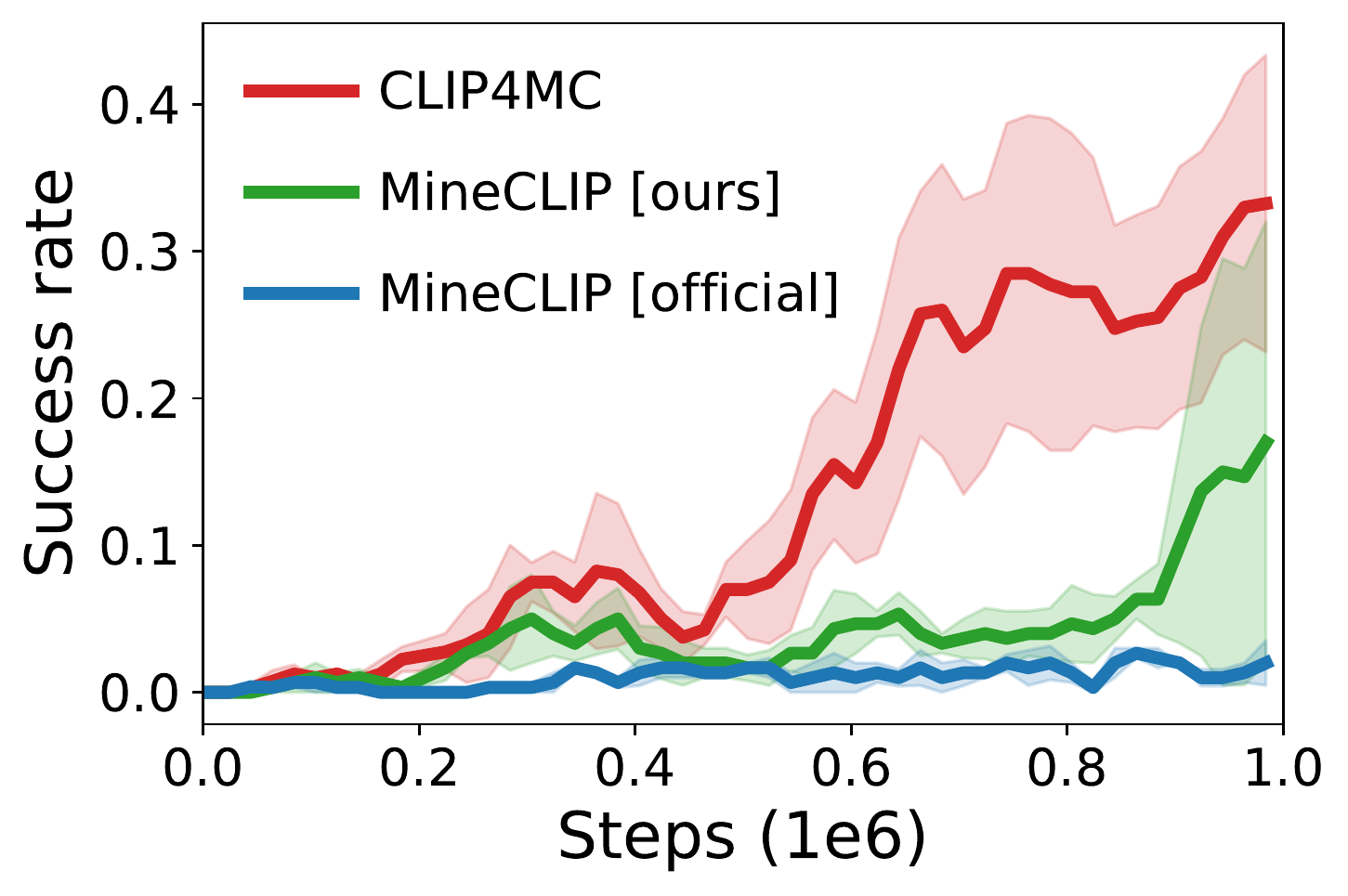}
		\caption{hunt a pig}
    \end{subfigure}
    \hspace{-0.1in}
    \begin{subfigure}{0.25\textwidth}
		\centering
		\includegraphics[width=\textwidth]{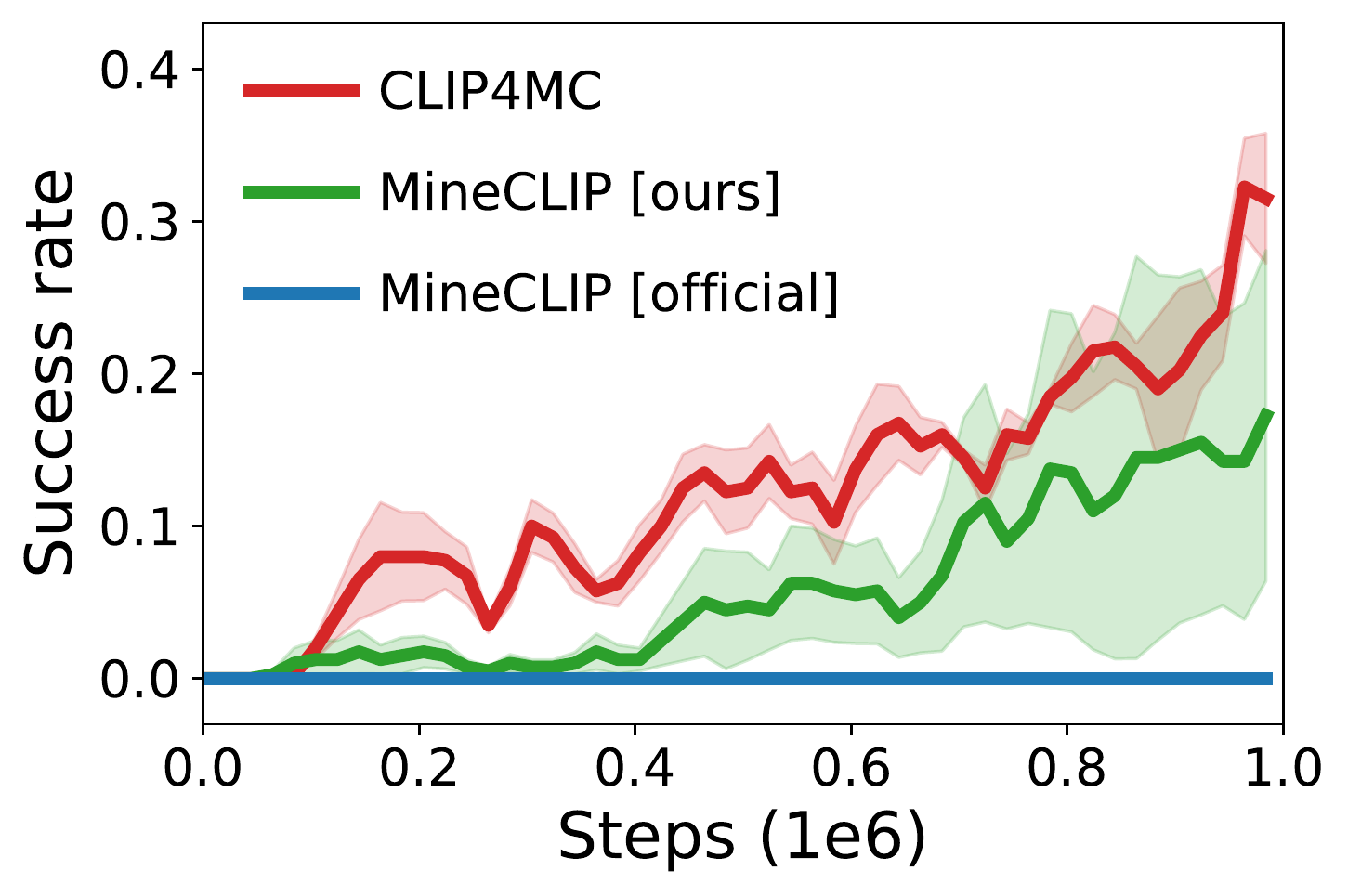}
		\caption{hunt a chicken}
    \end{subfigure}
    \caption{Learning curves of CLIP4MC, MineCLIP [ours], and MineCLIP [official] on eight Programmatic tasks of MineDojo.}
    \label{fig:result}
    \vspace{-2mm}
\end{figure}

\Cref{fig:result} shows the learning curves of all methods on eight Minecraft tasks. The results in \Cref{tab:mt_combat} are averaged on the last five checkpoints in \Cref{fig:result}.

\section{Creative Tasks}
\label{app:creative}

As defined in MineDojo \cite{fan2022minedojo}, Creative Tasks is a unique task suite, distinct from Programmatic Tasks due to the lack of success criteria. MineCLIP demonstrates its effectiveness in learning Creative Tasks, such as \textit{dig a hole} and \textit{lay the carpet}, and serving as a reliable success classifier. It is noticeable that Creative Tasks often lack clear indicators of task completion progress, unlike Programmatic Tasks. In other words, our adopted approach, where the size of the key entity serves as a surrogate for task completion progress, is ineffective. For example, in \textit{dig a hole}, it is not feasible to detect a hole and estimate its size before the agent completes this task. Therefore, the swap operation in CLIP4MC would bring no benefit to Creative Tasks. However, due to the upper bound of swap probability $P_{\rm max} = 0.5$ we set, CLIP4MC is expected to preserve the ability of MineCLIP to understand these creative behaviors without an explicit target entity that exists in the Minecraft world in advance.

\begin{figure}[h]
    \centering
    \includegraphics[width=0.5\textwidth]{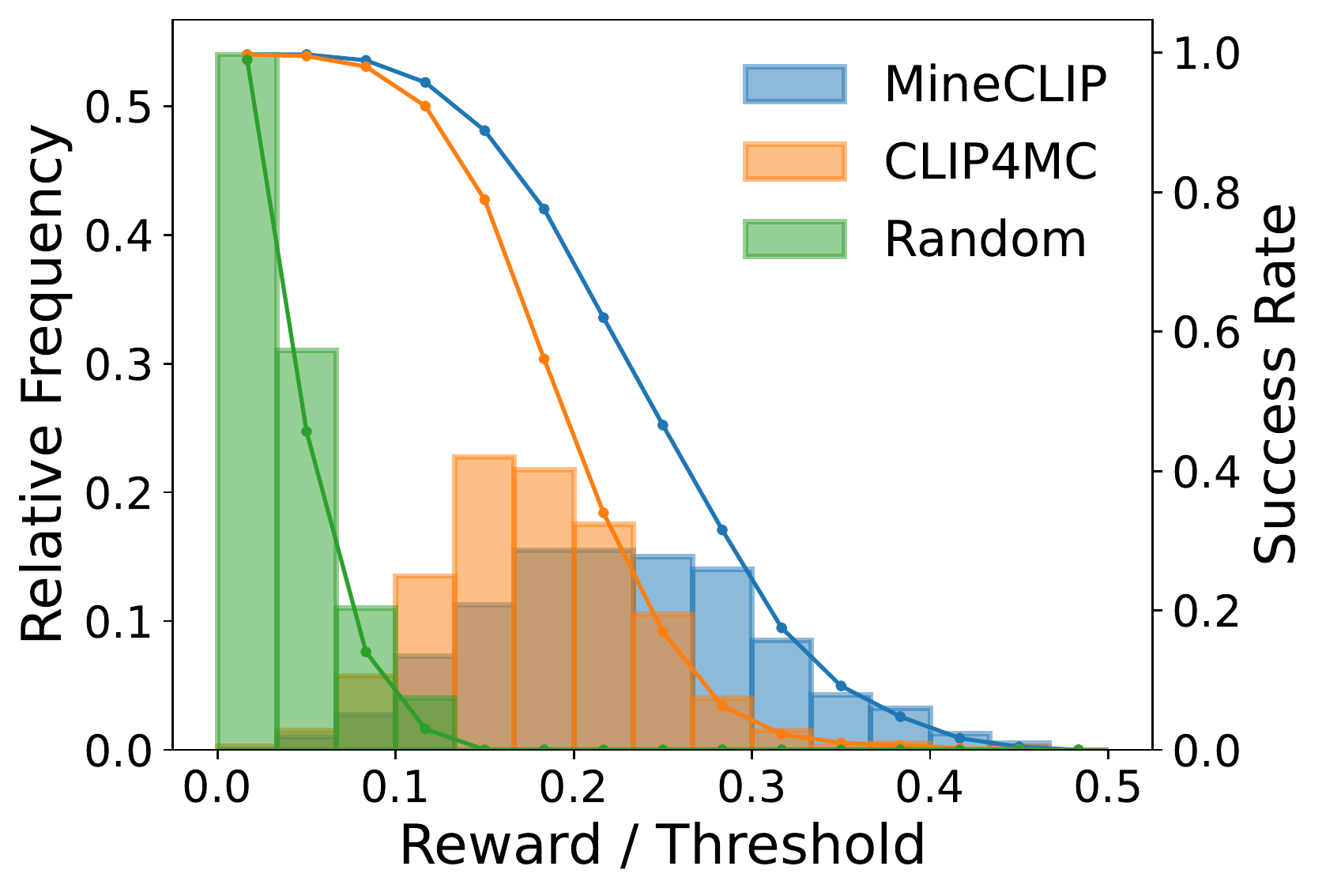}
    \caption{\textbf{Histogram \& left y-axis.} The distributions of MineCLIP reward for agents trained with CLIP4MC and MineCLIP reward and the agent taking random actions. \textbf{Curve \& right y-axis.} The success rates of agents with different reward thresholds.}
    \label{fig:dig}
\end{figure}

To validate our hypothesis regarding the effectiveness of CLIP4MC in Creative Tasks, we conduct experiments on \textit{dig a hole}. We train agents using PPO with intrinsic rewards calculated by CLIP4MC or MineCLIP [official]. At the beginning of each episode, the agent is spawned in the biome \texttt{plains} with an iron shovel. The prompt is ``dig a hole'' and each episode lasts 200 steps. Due to the lack of success criteria for \textit{dig a hole}, we run each agent for 100 episodes and record the average reward per step within each episode, calculated by MineCLIP [official]. The resulting reward distributions are shown in \Cref{fig:dig}. The result shows that the mean MineCLIP reward of the agent trained with CLIP4MC reward is lower than that of the agent trained with MineCLIP reward. This is consistent with expectations, as the agent trained with MineCLIP reward should overfit it and thus achieve higher rewards. In addition, the MineCLIP reward of the agent trained with CLIP4MC is significantly higher than that taking random actions, indicating that CLIP4MC successfully guides agents in learning behaviors relevant to \textit{dig a hole}.

The original MineCLIP paper proposes a successful classifier based on the average MineCLIP reward. In detail, a trajectory with a average MineCLIP reward greater than the threshold $\tau$ is classified as successful. However, the specific threshold values for each task are not provided. Therefore, we evaluate success rates at various reward thresholds, as depicted in \Cref{fig:dig}. Considering the high success rate (91.6\%) reported for \textit{dig a hole} in the original paper, we speculate that the agent trained with CLIP4MC would also achieve high success rates. This is supported by the narrow performance gap between the agent trained with MineCLIP and CLIP4MC rewards when the former's success rate is high (around 90\%), as shown in \Cref{fig:dig}.

\end{document}